# Can AI Debias the News? LLM Interventions Improve Cross-Partisan Receptivity but LLMs Overestimate Their Own Effectiveness

Faisal Feroz[1, 2, 3] & Jonas R. Kunst[4]

1. Department of Experimental Psychology, University of Oxford
2. Uehiro Oxford Institute, University of Oxford
3. Psychology Programme, School of Social Sciences, Nanyang Technological University
4. Department of Communication and Culture, BI Norwegian Business School

Corresponding author: Jonas R. Kunst, Department of Communication and Culture, BI Norwegian Business School. Email: jonas.r.kunst@bi.no

**Abstract**

Partisan news media erode cross-partisan trust, but large language models (LLMs) offer a potential means of debiasing such content at scale. Across two pre-registered experiments, we tested whether LLM-generated debiasing of liberal news headlines could improve conservative readers' trust-relevant judgments. Study 1 found that subtle lexical debiasing (replacing emotive words with moderate synonyms) had no effect on any outcome. Study 2 found that a more substantive reframing intervention significantly increased conservatives' perceived trustworthiness, completeness, and willingness to engage with liberal news headlines, without producing a backfire effect among liberals. In Study 1, the intervention produced robust effects across silicon participants simulated with six different models (o3-mini, o3, GPT-4o mini, GPT-4o, GPT-5 mini, and GPT-5), whereas it had no impact on human readers. In Study 2, the intervention's effects among silicon participants generally aligned directionally with human responses but were significantly larger for some outcomes, and three models (o3, GPT-4o mini, and GPT-4o) incorrectly predicted a liberal backfire effect absent in humans. Moderation analyses revealed that the models' implicit theory of who responds to debiasing diverged from the psychological profile that actually predicted human responsiveness. Most strikingly, in Study 2, participants simulated by each of the six models suggested that debiasing effects would be stronger among participants high in political in-group identification. Yet, no such moderation was observed among human participants. Collectively, these findings demonstrate that LLM-based debiasing can improve cross-partisan receptivity when targeting ideological framing rather than surface-level language, but that current models lack both the quantitative accuracy and qualitative psychological fidelity to evaluate their own interventions without human oversight.

## Introduction

The contemporary media landscape has undergone a structural transformation defined by increasing polarization and fragmentation. As news organizations cater to distinct ideological niches, the information ecosystem has bifurcated, often presenting fundamentally incompatible narratives of the same events (Cramer, 2011). Crucially, this divergence is reinforced by an increasing reliance on moralized and emotionally charged rhetoric. While such language successfully galvanizes partisan audiences, it signals ideological allegiance that actively decreases receptivity among political outgroups (Bentivegna & Rega, 2024). This supply-side polarization erodes the shared understanding necessary for democratic deliberation. Recent advances in large language models (LLMs) offer novel possibilities for mitigating this fragmentation at scale, yet it remains an open question whether algorithmic interventions can effectively bridge these divides or if they inevitably collide with the defensive processing mechanisms of news consumers.

Here, we report two pre-registered experimental studies testing whether LLM-generated debiasing can shift trust-relevant judgments and information receptivity among ideologically opposed readers. We further test whether LLMs accurately predict the effects of their own interventions, providing critical information about the viability of autonomous debiasing systems versus the need for humans in the loop. Specifically, Study 1 examines whether minimal lexical debiasing improves conservatives' receptivity to content from a liberal news source. Study 2 tests the effect of a more substantive reframing intervention, while also assessing reactions among liberals, allowing us to test for asymmetric and backfire effects. Critically, both studies test three potential moderators: strength of political in-group identification, general media trust, and cognitive flexibility.

**Partisan Bias Is Increasing in News Media**

Today's partisan trust gap reflects a dramatic transformation that has unfolded over recent decades. Longitudinal analyses of cable news reveal that major networks have grown increasingly distant from one another ideologically, a divergence that accelerated markedly after the 2016 US presidential election (Martin & Yurukoglu, 2017; Kim et al., 2022). While news audience polarisation has emerged across democracies, comparative research across twelve countries finds it is highest in the United States (Fletcher et al., 2020). The rise of partisan media has been accompanied by (and contributes to) broader affective polarisation, as citizens increasingly view political opponents not merely as misguided but as immoral and threatening to the nation's wellbeing (Pew Research, 2014, 2016). Affective polarisation refers to the tendency for partisans to dislike and distrust those on the other side of the political divide, independent of policy disagreements (Iyengar et al., 2019). Out-group animosity exceeds in-group loyalty as a driver of political behavior (Finkel et al., 2020), and moralized content spreads more readily within partisan networks than across them (Brady et al., 2017).

This trend poses a serious challenge to democratic discourse. When partisans consume news exclusively from ideologically aligned sources, they encounter different facts, priorities, and diverging interpretive frames, eroding the shared informational foundation that deliberative democracy requires. Moreover, exposure to partisan content deepens hostility toward out-groups, which in turn can reduce collaboration and compromise across partisan lines (Garrett et al., 2014). These dynamics underscore the need to identify strategies that can make partisans more receptive to information presented from the other side's perspective.

**The Potential of Using AI to "Debias" News Media**

Biased language and framing directly impacts how readers receive and trust information. Linguistic markers of partisan slant polarise outgroups while building reputational trust amongst ingroups (Walker et al., 2025). When encountering language that signals out-group authorship, partisans typically discount the credibility of factual claims and resist persuasive appeals they might otherwise find compelling. As a result, the style of writing itself may trigger defensive processing before readers substantively evaluate what they are reading.

Prior research has explored whether reducing linguistic bias can improve receptivity and trust. Studies manipulating the partisan valence of news framing have shown that neutralizing loaded language can increase perceived fairness among opposing partisans (Walker et al., 2025). Similarly, work on "bridging" interventions demonstrates that reframing political arguments in terms of the opposing side's moral values can enhance cross-partisan persuasion (Voelkel & Feinberg, 2017; Feinberg & Willer, 2019). While these interventions are promising, human debiasing efforts face a fundamental scalability bottleneck: professional editors cannot feasibly review and revise the vast volume of content produced daily across thousands of outlets.

The advent of large language models potentially changes this calculus. Kuo et al. (2025) demonstrated that GPT-4o Mini achieved 92.5% agreement with human annotators in identifying biased paragraphs and can iteratively revise content to reduce detected bias while preserving meaning. Human raters judged the debiased outputs as less biased than originals, suggesting that LLM interventions can shift at least some perceptions of slant. However, Baris Schlicht et al. (2024) found that conversational LLMs frequently introduced factual errors, altered contextual meaning, and made unnecessary stylistic changes that diverged from expert editorial judgment, raising concerns about deployment in real newsrooms.

Critically, to the best of our knowledge, existing work has no*t* tested whether debiasing influences the trust-relevant judgments that matter most for cross-partisan engagement. Perceived bias and perceived trustworthiness are related but distinct constructs: readers may acknowledge that an article uses neutral language while still distrusting it. Furthermore, to our knowledge, no prior work has examined whether debiasing influences readers' willingness to engage with the perspectives an article presents, which is arguably the outcome most consequential for bridging partisan divides, and to what extent psychological differences among receivers modulate the effectiveness of such debiasing.

**Potential Psychological Factors Hindering or Facilitating the Efficacy of Debiasing**

Except for considering political affiliation as a characteristic, previous debiasing research has mostly treated partisan audiences as monolithic groups who respond uniformly to interventions. This approach obscures potentially consequential heterogeneity in the population. We next outline three theoretically and empirically derived psychological variables for which theory and previous work suggest various, sometimes opposing, influences on partisans' responsiveness to debiasing.

First, baseline media trust may shape receptivity to debiasing interventions. Individuals who distrust media institutions generally may approach all content with heightened skepticism (Park et al., 2024), rendering surface-level linguistic changes insufficient to shift their judgments. Alternatively, these highly skeptical readers may experience a positive violation of expectations (for a review, see Margoni et al., 2024): precisely because they anticipate partisan slant, algorithmically neutralized content may stand out as surprisingly neutral, rendering high-distrust readers more responsive to the intervention.

Second, cognitive flexibility – the capacity to update beliefs and shift perspectives in response to new information (Uddin, 2021) – may determine whether individuals revise initial impressions triggered by linguistic framing. Debiasing may be more effective among those with high cognitive flexibility, who are less rigid in their information processing. However, a plausible opposing prediction exists: highly flexible individuals may be adept at motivated reasoning, using their cognitive resources to detect subtle ideological cues or construct sophisticated justifications for rejecting the debiased content (Kahan & Corbin, 2016), ironically dampening the intervention's effect.

Third, and arguably most centrally, the strength of partisan identification may moderate defensive processing. The hostile media effect, which is the tendency for partisans to perceive neutral or balanced coverage as biased against their own position, has proven remarkably robust across topics, media formats, and national contexts (Vallone et al., 1985; Hansen & Kim, 2011), and may be amplified among strong identifiers (Reid, 2012). Strong identifiers may be most anchored in their beliefs, making them less affected by out-group-aligned information regardless of its linguistic tone and framing. At the same time, since strong identifiers are vigilant for signs of out-group bias (Lilienfeld & Latzman, 2014), they may alternatively respond most favorably to linguistic changes that attenuate such threats.

Fourth, and finally, it is critical to consider the potential costs of debiasing among the outlet's primary, ideologically aligned audience. While interventions aim to increase receptivity among the out-group, they risk triggering a backfire effect among the in-group by diluting the identity signals they expect. Partisan media consumers often seek validation rather than neutrality; they value content that aggressively defends their worldview and signals shared moral commitments (Finkel et al., 2020). Consequently, an MSNBC article stripped of its liberal framing may be perceived by liberal readers not as "objective," but as "watered down" or engaging in false equivalence: a successful intervention for one group may constitute a degradation of quality for the other.

**The Need to Assess LLM Predictions of Debiasing Efficacy**

If the goal is to develop scalable debiasing systems, it is critical to assess whether the models behind such systems have a realistic sense of how effective their debiasing is—that is, whether their implicit model of human psychology accurately predicts how real partisans will respond. A persistent criticism of large language models is that they lack genuine world models and instead rely on statistical patterns that may diverge from causal mechanisms governing human cognition (Bender et al., 2021; Gopnik, 2023; Mitchell & Krakauer, 2023). This limitation becomes consequential when models are deployed to influence human judgments: an LLM may confidently "debias" content in ways that satisfy algorithmic criteria while leaving human perceptions unchanged.

Evidence from adjacent domains suggests some grounds for concern. Research on LLM-human alignment reveals that models' implicit theories of what resonates with specific audiences often diverge from human reality. For instance, Hackenburg et al. (2025a) found that microtargeting LLM-generated political messages produced only small additional effects on persuasion (under one percentage point) and Hackenburg et al. (2025b) found that partisan role-playing conferred little advantage—participants could identify an LLM's intended partisan identity only 46% of the time. Related limitations appear in psychological and behavioral domains. While a scoping review of LLM applications in mental health care noted some promise, its authors concluded that current evidence is insufficient to support the use of LLMs as standalone interventions (Hua et al., 2025). More generally, while some models can approximate expert judgments in behavioral sciences and psychiatric diagnoses, performance varies sharply across models and tasks (Lippert et al., 2024; Gargari et al., 2024). Together, these findings suggest that LLMs may possess (promising yet) incomplete or distorted representations of the features that determine how interventions are received by the populations they aim to influence.

Two approaches exist for assessing LLM predictions of intervention efficacy. The first, simpler approach involves directly prompting models to predict how human participants will respond to experimental conditions (Almeida et al., 2024, Lippert et al., 2024, Mei et al., 2024). The second, arguably, more sophisticated approach employs silicon participants – LLM instances prompted with demographic profiles matching target populations – to fully simulate studies and compare model-generated response patterns against human data (Argyle et al., 2023; Cui et al., 2025; Wedyan et al., 2025). In this paradigm, the model is assigned a specific persona (e.g., "a conservative 40 year old male voter") and asked to complete the survey tasks from that perspective, allowing researchers to administer identical evaluation tasks to models and humans, revealing systematic discrepancies in how each responds to interventions. In a replication of 156 psychology experiments using GPT-4, Claude and Deepseek, Cui, Li and Zhou (2025) found that the LLMs replicated main effects with large success (73–81%) and interaction effects (46–63%) with moderate success; however, they generally overinflated effect sizes. Given the stakes involved in deploying AI-assisted editorial tools at scale, testing whether LLMs accurately anticipate the effects of their own debiasing interventions is essential, particularly in a domain where misalignment could mean optimizing for features that lead to further mistrust, misunderstanding and polarisation.

**The Present Research**

Integrating research on partisan media bias, the hostile media effect, and emerging work on LLM-based text interventions, the present research investigated whether algorithmically debiasing partisan news content can shift trust-relevant judgments and information receptivity among ideologically opposed readers. Whereas prior work has demonstrated that LLMs can reduce detected linguistic bias in news articles (Kuo et al., 2025; Baris Schlicht et al., 2024), existing research has no*t* tested whether such debiasing influences the trust-relevant judgments and willingness to engage with opposing perspectives, and has largely treated partisan audiences as monolithic. The present work addresses these gaps.

Our research consisted of two pre-registered randomized experiments that progressively escalated the depth of the debiasing intervention. Study 1 tested whether minimal lexical debiasing, specifically the targeted replacement of emotive or moralized language while preserving narrative structure, is sufficient to shift conservative participants' trust-relevant judgments of liberal news content. Study 2 extended this by testing a more substantive reframing intervention that altered not just word choice but the framing of the issue itself. Critically, Study 2 also included liberal participants, allowing us to examine whether debiasing effects are symmetrical across the political spectrum or whether a backfire effect occurs among the ideologically aligned in-group audience. In both studies, we modelled three psychological moderators, namely baseline media trust, cognitive flexibility, and strength of partisan identification, to identify boundary conditions and facilitating factors of algorithmic debiasing.

Both studies used opinion headlines and subheadings as stimuli. This focus is motivated by the outsized role headlines play in contemporary news consumption: headlines are the primary cue on which readers decide whether to continue reading (Dor, 2003). Experimental work has shown that headlines shape readers' memory, inferential reasoning, and behavioral intentions even when the article body is read in full (Ecker et al., 2014). In digital and social media environments, the primacy of headlines is even more pronounced. An analysis of sharing behavior on Twitter estimated that 59% of shared links were never clicked (Gabielkov et al., 2016), and a large-scale study of over 35 million Facebook posts found that approximately 75% of news links were forwarded without the sharer first clicking on them (Sundar et al., 2024). Headlines thus represent the most consequential unit of news text for many readers, making them a high-priority target for debiasing interventions.

In addition to testing the effects of debiasing on human participants, both studies incorporated a silicon participant methodology (Argyle et al., 2023; Cui et al., 2025; Wedyan et al., 2025) in which LLM instances prompted with demographic profiles exactly matching the human samples completed the same experimental tasks. This comparison allowed us to quantify the alignment gap: the discrepancy between how an LLM *predicts* a conservative will react to debiasing versus how an actual conservative responds. This provides the first critical test of whether current models possess a sufficiently accurate theory of mind to serve as autonomous debiasing tools.

Theoretically, these studies adjudicate between competing accounts of the hostile media effect (Vallone et al., 1985). If linguistic debiasing significantly improves receptivity, it suggests that perceived bias is largely stimulus-driven (Caparos et al., 2015), that is, a reaction to specific textual cues that can be engineered away. Conversely, if interventions fail despite objective neutralization, it supports the view that perceived bias is primarily a top-down projection of identity (see, e.g., Mohanty, et al., 2025), impervious to textual refinement. Methodologically, our comparison of human and silicon participants provides a rigorous audit of LLMs as tools for social science simulation in the domain of political polarization. By determining whether synthetic data accurately captures the nuance of defensive processing or merely stereotypes partisan behavior, we clarify the validity of using AI agents to pilot social interventions.

Practically, as newsrooms and platforms consider integrating generative AI to diversify audiences or depoliticize content (Cools & Diakopoulos, 2024), it is vital to understand the risks of automation. If models predict efficacy where none exists, or if they introduce changes that backfire among readers who do not share the ideological assumptions embedded in the intervention, blind deployment could exacerbate the very polarization it aims to solve. Our results thus inform the necessary human-in-the-loop protocols (Mosqueira-Rey et al., 2022) required to deploy AI responsibly in the epistemic commons.

## Study 1

Study 1 provided an initial test of whether minimal lexical debiasing is sufficient to shift trust-relevant judgments among partisan readers. We focused on a deliberately stringent test of receptivity: conservative participants evaluating opinion headlines from MSNBC, a liberal outlet toward which conservatives typically show low trust (Pew Research, 2024).

In a within-subjects design, participants read ten MSNBC opinion headlines that had been pre-tested as the least trustworthy from an initial pool of twenty according to a conservative pilot sample. Each headline was randomly presented either in its original format or in a version that had been lexically debiased by an LLM, in which emotive and moralized language was automatically replaced while preserving the original narrative structure and factual content. After reading each headline, participants rated its perceived trustworthiness, the extent to which it told the whole story, and their willingness to consider the perspective presented.

We predicted that debiased headlines would elicit higher ratings of trustworthiness, higher perceptions of the article telling the whole story, and greater willingness to consider the presented perspective compared with original headlines. We further explored whether the effect of debiasing would be moderated by three individual-difference variables: baseline (i.e., trait-like) media trust, cognitive flexibility, and strength of conservative identification. As outlined in the introduction, theory and prior work suggest competing predictions for each moderator, and we therefore treated these interactions as exploratory.

In addition to testing human participants, we ran a parallel silicon participant procedure in which LLM instances assigned conservative demographic profiles completed the identical experimental task. By comparing the simulated response patterns to those of actual conservative participants, we obtained an estimate of the potential discrepancy between LLM predictions and human psychological reality in this domain.

## Method

### *Human Participants*

Based on a Monte Carlo power simulation, 170 participants were needed to provide 90% chance to observe two-way cross-level interaction (manipulation × continuous moderator) as specified in our pre-registration (https://osf.io/nxaqf/overview?view_only=b662a48107b64c0f92d048a3e238d64f), for an effect size of Cohen's $d$ = 0.25 at a significance criterion of 0.05. As we tested a novel procedure, this small effect size was selected. Therefore, 179 Republican US participants were recruited from Prolific for our experiment. Participants saw an experiment titled “Study about news articles” and were informed that they would be compensated £1.50 for their time. The experiment took 13 minutes on average. To guard against AI-agent responses in our experiments, we limited our submissions to Prolific participants with an approval rate between 99-100%, with between 50 to 10,000 completed tasks. We also set the recruitment to be gender balanced – 50% men and 50% women.

Participants indicated that they were above 18 years of age and spoke English fluently (requirements for participating). Of the 179 participants, 49.2% identified as women and 50.8% as men. In terms of education, 5.0% had completed high school or equivalent, 7.3% had some college experience without obtaining a degree, 5.6% held an associate or technical degree, 39.1% held a bachelor's degree, and 43.0% held a graduate or professional degree (e.g., MA, PhD, JD, MD). Regarding racial identification, 68.2% identified as White, 29.6% as Black or African American, 2.2% as Asian, and 0.6% as American Indian or Alaska Native; 0.6% preferred not to disclose their race. Some participants selected multiple racial categories, including 0.6% who identified as both White and Asian, and another 0.6% as both White and Black or African American.

Three participants failed two out of three comprehension checks. Exclusions based on this pre-registered requirement left us with 176 participants who completed 1,760 trials in total.

***Silicon Participants***

To estimate how effective the LLM expected the debiasing procedure to be, we replicated the study, replacing the Republican human participants with simulated "silicon" participants generated using OpenAI's o3-mini model (see Cui, Li and Zhou, 2024 and Wedyan et al., 2025 for other examples of using silicon participants). A total of 176 simulated participants were created, matching the post-exclusion human analytic sample exactly. Each GPT participant was assigned the demographic profile – age, gender, and race, scores on the moderators Media Trust, Cognitive Flexibility, and In-group Identity, as well as the familiarity the participant had with each news headline – of a corresponding human participant from Study 1a, ensuring a one-to-one mapping between human and silicon participants. Aligned with standards in the field (Argyle et al., 2023), these attributes were embedded directly into the GPT prompt to encourage the model to adopt a perspective consistent with the lived experience of a U.S. Republican participant. All prompts followed the same structure as in the human task, and no temperature parameter was passed in any call: o3-mini is a reasoning model whose API does not accept one, so the model's default sampling applied throughout (see the supplementary online materials here: https://osf.io/na78b/overview?view_only=8503c0c0d91649db80f903e05d5ba5b3 for the exact prompts used).

Simulations were run using a parallelized Python script with retry logic for rate limits and formatting corrections. GPT was required to "read" 10 articles per prompt and return responses in a structured JSON format, with each API call corresponding to a single simulated participant. Responses that failed JSON validation were re-queried until a valid response was obtained, yielding 1,760 valid trials (176 participants x 10 trials).The full codebase and raw outputs are available in the SOM.

### ***Procedure***

Both studies presented in this paper were approved by the ethical review board of the institution of the senior author (Nr. pNr051). The current study was pre-registered at https://osf.io/nxaqf/overview?view_only=b662a48107b64c0f92d048a3e238d64f). Partisan news articles (comprising a title and a subheading line) for our experiment were obtained from the Opinions section of MSNBC News (url: https://www.msnbc.com/opinion/columnists) on 26th April 2025. MSNBC was selected because it is consistently rated as a left-leaning news outlet by media bias rating organizations (Ad Fontes Media, 2024; AllSides, 2024), making it an appropriate source for testing effects of debiasing among conservative participants. Selenium (version 4.31.0, Selenium Team, 2025) and BeautifulSoup (version 4.13.3, Crummy, 2025) packages were used in a Python script to scrape the newest articles, after which the 20 latest articles relevant to US news were identified. To identify the most polarizing articles for use in the main experiment, we pretested the news opinion headlines with a different sample of conservative participants ($N$ = 121). Participants rated each headline's trustworthiness on a 7-point Likert scale (1 = *Strongly Disagree* to 7 = *Strongly Agree*) in response to the statement "I think the headline is trustworthy" (adopted from Tsfati et al., 2023). Based on these ratings, we selected the ten headlines with the lowest average trustworthiness, ranging from $M$ = 3.38 ($SD$ = 1.77) to $M$ = 4.01 ($SD$ = 1.62), for inclusion in the main study. This selection strategy was deliberately designed to yield maximally divisive yet ecologically valid stimuli, maximizing the potential impact of debiasing interventions while maintaining baseline relevance and recognizability.

These ten articles went through a "debiasing" process through the use of ChatGPT o3-mini. We selected OpenAI's o3-mini (released January 31, 2025) as the primary model for generating the debiasing interventions and simulating the silicon participants, based on its balance of reasoning quality and cost-efficiency at the time of data collection. It was reported to match the larger o1 model's performance while producing responses substantially faster (OpenAI, 2025), and at roughly $1.10 and $4.40 per million input and output tokens, it was priced at a fraction of contemporaneous flagship models such as o1 (approximately $15 and $60, respectively). Using the ChatGPT API, articles were individually presented to ChatGPT o3-mini, alongside instructions to "...replace emotionally charged words with more moderate synonyms while preserving facts, viewpoint, and grammatical structure…" The goal of this approach was minimal invasiveness: to reduce perceived bias through subtle lexical changes without altering the article's core argument or factual content. Study 1's prompt thus constrained the model to word substitution: each replacement had to be a single word or short phrase of lower intensity but the same semantic polarity, with capitalization, punctuation, and tense preserved (e.g., its worked example "kick" → "remove"; in the stimuli, "attack" → "criticism"). Examples of sample and debiased articles (not included in the actual stimuli) were also provided to the model in the API request. Table 1 presents the original stimuli headlines alongside their Study 1 (lexical) and Study 2 (reframing) debiased versions. For the full prompt and exact script, please refer to the supplementary material (https://osf.io/na78b/overview?view_only=8503c0c0d91649db80f903e05d5ba5b3).

To verify that the debiasing intervention preserved meaning, an independent sample of Republican/conservative raters (*N* = 87) viewed each original/debiased headline pair side by side and rated whether the two versions convey the same core information (1 = *completely different meaning*, 7 = *identical meaning*) and whether any factual claims were added, removed, or changed (Yes/No). Ratings fell toward the identical-meaning pole ($M = 5.74$, $SD = 1.53$), and only 24.0% of judgments indicated a change in factual claims.

A separate sample of 100 Republican/conservative raters viewed both versions of each Study 1 article side by side and rated each on perceived anti-conservative bias (0–100). Debiased versions were rated significantly less biased ($M = 59.4$ vs. 63.1; $b = -3.74$, $SE = 0.89$, $p < .001$, $d_z = 0.37$). The validation raters additionally judged one randomly chosen version of each headline in isolation on how likely they would be to come across it while scrolling through news online (1 = very unlikely, 7 = very likely); original and debiased versions did not differ ($M = 5.03$ vs. 5.06, $p = .692$), indicating that the rewrites read as naturally occurring headlines.

In the experiment, human and silicon participants then underwent ten trials. In each trial, they were presented first with an opinion article, randomised to either be original or debiased by o3-mini (see SOM for articles and an example trial; silicon participants were assigned to the exact same conditions as their matched human counterparts). After reading the article, participants were asked the extent they agreed or disagreed with four dependent or control variables: topic familiarity (control), headline trustworthiness, perceived comprehensiveness, and openness to consider the article's perspective. As pre-registered, topic familiarity was included as a control variable because prior exposure to or knowledge of an issue may influence how participants evaluate the article's trustworthiness and bias, independent of the debiasing manipulation. Data collection was conducted in a timely manner: the study was deployed on Prolific on 26 April 2025, the same day the articles were scraped, and all 176 responses were collected within approximately 16 hours (ending 01:44 UTC on 27 April 2025). This ensured the stimuli remained current and recognisable as recent news at the time of rating. After the ten trials, participants were presented with the Media Trust scale (Meyer, 1988), Cognitive Flexibility scale (CFS; Martin & Rubin, 1995), In-Group Identification Measure (IGI; Leach et al., 2008) and questions about their demographics (education, age, race, gender). At the end, participants were presented with the ten articles again and asked the extent to which each article shows anti-conservative bias (our manipulation check). We report all data, experimental conditions, and instruments.

**Prompt Engineering.**

In both studies, debiasing prompts were engineered as single-turn, zero-shot instructions to o3-mini (o3-mini-2025-01-31), submitted as a plain user message with no system prompt and default decoding parameters. Each prompt followed the same fixed four-part structure: (a) a TASK statement defining the rewriting objective; (b) eight numbered RULES constraining the edit space; (c) a FORMAT specification ("Debiased Headline:" / "Debiased Description:") that allowed deterministic parsing of the output; and (d) three worked EXAMPLES (original–debiased pairs) illustrating the intended edit granularity, followed by the input item. Each of the 10 articles was rewritten in a separate, independent API call, so no article's rewrite could influence another's.

The two prompts were developed sequentially: the Study 2 prompt was created by editing the Study 1 prompt, holding constant every constraint that protects fidelity to the source while relaxing a single constraint governing edit depth. Specifically, both prompts share four invariant rules: (1) no factual content may be added, deleted, or reordered; (2) legitimate criticism and negative judgments must be retained, with only their emotional intensity reduced; (3) edits are limited to a maximum of two per sentence unless strictly required; and (4) the valence of any described outcome must remain unambiguous (e.g., "negatively impact" rather than "impact"). The prompts differ in the operation they license. Study 1 restricted the model to lexical substitution: its job was "ONLY to replace emotionally charged words with more moderate synonyms while preserving facts, viewpoint, and grammatical structure," each replacement being a single word or short phrase of lower intensity but the same semantic polarity, with capitalization, punctuation, and tense preserved. Study 2 broadened the objective to rephrasing "in a way that lowers emotional intensity and aligns with a conservative communication style," explicitly permitting sentence-level restructuring and passive constructions "if this helps reduce perceived bias." The worked examples were rewritten in parallel to model the deeper edit. The same three source headlines that Study 1's examples resolve through word swaps (kick → remove; lawless → unlawful; incorrect → inaccurate), Study 2's examples resolve through restructuring and attribution ("None of us are safe from a lawless White House" → "Concerns grow over unlawful actions by the current White House"). The prompts were thus designed so that Studies 1 and 2 differ by one engineered degree of freedom (whether the model may alter rhetorical framing), enabling the contribution of framing beyond vocabulary to be isolated.

Outputs were parsed programmatically from the labeled format and merged with the originals into the published stimulus files; the saved generator outputs match the published stimuli verbatim. The prompt templates themselves were developed through iterative refinement in discussion between the authors. Candidate wordings of the task statement, rules, and worked examples were drafted, the resulting rewrites inspected against the fidelity constraints (fact preservation, retained criticism, unambiguous valence), and the prompt revised until the generated stimuli satisfied the intended manipulation, as indicated by surface-level lexical softening in Study 1 and audience-tailored reframing in Study 2. Because the API calls set no random seed and o3-mini is nondeterministic, re-running the generators reproduces the procedure but not the exact published wording; the complete final prompt templates and generation code are provided in the Supplementary Materials.

**Table 1**

*Example Item Illustrating the Two Debiasing Interventions: Lexical Substitution (Study 1) Versus Substantive Reframing (Study 2)*

| Version | Headline | Description (teaser) | What was changed | Method |
|---|---|---|---|---|
| *Original (MSNBC)* | Trump's new **attack** on the Civil Rights Act is a **license** to discriminate | In a new executive order, the White House is **targeting** making it easier for businesses to discriminate during the hiring process. | — (source text; affectively charged terms marked in red) | Scraped verbatim from the MSNBC opinion-columnists section (headline + teaser); politically filtered set of 10 articles (*msnbc_articles_politically_filtered.json*), identical input to both generators. |
| *Debiased, Study 1* | Trump's new **criticism of** the Civil Rights Act is a **permission** to discriminate | In a new executive order, the White House is **aiming at** making it easier for businesses to discriminate during the hiring process. | *Lexical only.* One-to-one synonym substitution of charged words (attack → criticism; license → permission; targeting → aiming at). Syntax, propositional content, and the author's evaluative stance are fully preserved. | o3-mini prompted to do "ONLY…replace emotionally charged words with more moderate synonyms while preserving facts, viewpoint, and grammatical structure": replacements must be a single word or short phrase of "lower intensity but same semantic polarity" and "STILL emotive"; max two edits per sentence; no deleting, adding, or reordering of facts; capitalization, punctuation, and tense preserved (*study1_generate_debiased_articles.py*). |

| *Debiased, Study 2* | Trump's **recent stance** on the Civil Rights Act **viewed as permitting** discrimination | In a new executive order, the White House **intends to simplify hiring regulations, a change that could permit** businesses to engage in discriminatory practices. | *Substantive (stylistic–ideological) reframing.* Sentence-level rewriting: the accusation becomes an attributed characterization ("viewed as"), certainty is hedged ("could permit"), and the policy's stated purpose ("simplify hiring regulations") is separated from its alleged consequence. | o3-mini prompted to "rephrase the headline and description in a way that lowers emotional intensity and aligns with a conservative communication style, while preserving all facts, core viewpoints, and grammatical structure": it "may rephrase sentences or introduce passive constructions if this helps reduce perceived bias"; charged words still replaced with "moderate but still evaluative alternatives"; no factual content added, removed, or reordered (*study2_generate_debiased_articles.py*). |
|---|---|---|---|---|

*Note.* Color coding: **red** = affectively charged language in the original; **blue** = Study 1 lexical substitutions; **green** = Study 2 substantive reframing.

### ***Measures***

Unless stated otherwise, all items were rated on a 7-point Likert scale (1 = *Strongly disagree*, 7 = *Strongly agree*).

**Dependent Variables**. To assess perceived trustworthiness of the headline, participants rated the statement "I think the headline is trustworthy". This item was adapted from the News Credibility Scale (Gaziano & McGrath, 1986; Meyer, 1988), widely used in prior research on media trust and perceived bias. Participants also rated the statement "I think the headline tells the whole story" on the same 7-point Likert scale. This item was likewise drawn from the News Credibility Scale, capturing perceptions of the article's comprehensiveness. Together, these two items operationalize perceived credibility, capturing whether readers trust the information and find it complete, as distinct from mere evaluations of political slant. As a measure of participants' receptiveness to the viewpoint presented in the opinion piece, we included the item "I would be open to considering the perspective in the opinion piece." This item was developed for the current study to assess willingness to engage with opposing or unfamiliar views. Finally, for each trial, participants also indicated their prior familiarity with the topic. This was measured with a single item: "I have seen discussions of this topic in the news recently."

**Moderators.** All moderator items were averaged when computing composite scores.

**Media trust** was assessed with the Media Trust Scale (Meyer, 1988), using nine items that measured participants' perceptions of the accuracy, fairness, and reliability of media outlets (e.g., "The media is unbiased"; $\alpha = .96$).

**Cognitive flexibility** was assessed using twelve items from Martin and Rubin (1995) capturing individuals' capacity to adapt to new information and generate alternative responses across situations (e.g., "I am willing to listen and consider alternatives for handling a problem"; $\alpha = .84$). Items 2, 3, 5, and 10 were reverse-scored prior to analysis.

**In-group identification** was measured using six items from Leach et al. (2008), assessing affective and identity-based connections to the group of conservative people (e.g., "The fact that I am conservative is an important part of my identity"; α = .93).

**Manipulation Check.** To assess whether participants perceived the political orientation of the articles as intended, we included a manipulation check at the end of the main task (i.e., presented in a new randomized loop). For each article, they were asked: "To what extent does the article show an anti-conservative bias?" Responses were recorded on a 7-point scale from 0 (Not at all) to 6 (Very much). This looped evaluation allowed us to measure perceived political bias independently of the main outcome measures and confirm whether the debiasing manipulation was effective. By administering this measure of perceived neutrality separately from the trial-by-trial credibility items, we were able to operationally distinguish readers' detection of partisan slant from their judgments of whether the headline was trustworthy or comprehensive.

**Attention Check.** Three attention-check items were embedded in the moderator scales to identify inattentive responding. Each item instructed participants to select a specific response option (e.g., "For this item, please select 7"). The three checks were placed within the Media Trust, Cognitive Flexibility, and In-Group Identification scales. As pre-registered, participants who failed two or more of the three attention checks were excluded from analyses.

***Data Analysis***

All analyses in this and the second study were conducted in R, using packages *lme4 version 1.1.37* (Bates et al., 2015) and *lmerTest version 3.1.3* (Kuznetsova et al., 2017). Our first set of analyses focused on the human sample. As pre-registered, we estimated mixed-effects models to account for the nested structure of the data, with trials nested within participants. Random intercepts and random slopes were included, provided that models converged. Analyses proceeded in two stages. First, we estimated models testing the main effect of the predictor of interest. Next, we extended these models by adding the moderators and their two-way interactions. In all analyses, we controlled for participants' prior familiarity with the topic. Separate models were estimated for each dependent variable.

In addition to the human sample analyses, we conducted the same models with responses from silicon participants generated by o3 mini. This constituted an exploratory test. For the replication with silicon participants, we used the original participants' corresponding values for individual differences scores recorded, such as familiarity, media trust, in-group identification, and cognitive flexibility. Finally, we run analyses pooling the data from the human and silicon participants to directly test for differences in effects. Throughout, β denotes semi-standardised coefficients from models estimated with jtools::summ(scale = TRUE), in which continuous predictors were standardised to $M = 0$ and $SD = 1$ while outcome variables remained on their original measurement scale (0–6 for the manipulation check; 1–7 otherwise). No coefficient is standardised with respect to the outcome; all βs are therefore expressed in raw scale points of the outcome. Coefficients of continuous predictors represent the change in the outcome associated with a 1-*SD* increase in the predictor. Coefficients of the categorical condition contrast, including the simple effects and simple slopes reported at levels of a moderator, are not rescaled by the standardisation and represent the difference between the debiased and original versions. Interaction coefficients involving condition represent the change in this difference, per 1-*SD* increase in a continuous moderator or between levels of a categorical moderator. For consistency, we use the symbol β for all coefficients from these models. Simple slopes were obtained via interactions::sim_slopes() with Satterthwaite degrees of freedom. All data, materials, and code can be obtained at https://osf.io/na78b/overview?view_only=8503c0c0d91649db80f903e05d5ba5b3.

## Results

### *Human Sample Results*

We conducted mixed-effects linear regression models predicting the effects of condition (original vs. debiased) on each outcome, controlling for topic familiarity, with random intercepts and slopes for condition by participant and random intercepts by trial (see Figure 1 for plots). Due to one human participant's partial response on a single trial, trial-level N ranges from 1,759 to 1,760 across models depending on the outcome and covariates included; exact N's can be read from the Satterthwaite degrees of freedom. All 176 human and 176 silicon participants contributed to every reported analysis.

**Figure 1**

*Mean Ratings by Conservative Participants for Original and Debiased Articles are Displayed for the Manipulation Check and the Three Dependent Variables Measured*

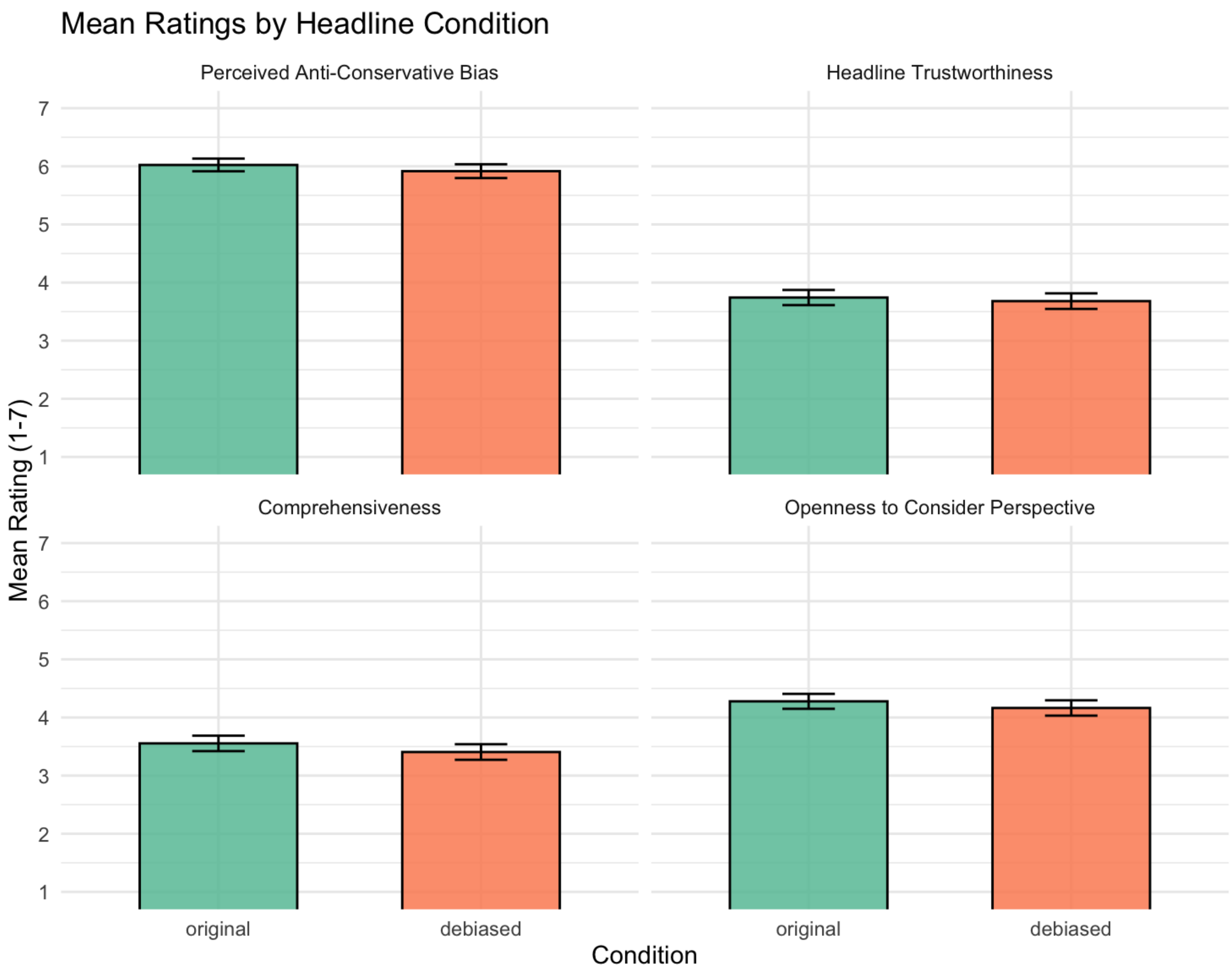

*Note.* Error bars represent 95% CI. Anti-conservative bias is measured on a scale from 0 to 6 and was transformed to 1 to 7 prior to visualization.

**Confirmatory Tests.** We first report the pre-registered tests of the effect of debiasing on each outcome.

***Anti-Conservative Bias (Manipulation Check).*** Contrary to expectations, the debiasing manipulation did not significantly reduce perceived anti-conservative bias, $\beta = -0.09$, 95% CI [-0.22, 0.04], $t(160.75) = -1.39$, $p = .167$. Topic familiarity was negatively associated with perceived bias, $\beta = -0.10$, 95% CI [-0.17, -0.02], $t(1657.22) = -2.40$, $p = .016$. To assess robustness, we also estimated this and all remaining models without the pre-registered covariate. Topic familiarity was often strongly associated with the outcome variables, so presenting both specifications allows readers to verify that the pattern of results does not depend on its inclusion. In this analysis, the effect of debiasing on perceived bias remained non-significant, $\beta = -0.09$, 95% CI [-0.23, 0.04], $t(161.92) = -1.40$, $p = .162$.

***Headline Trustworthiness.*** There was no significant effect of debiasing on perceived trustworthiness, $\beta = 0.05$, 95% CI [-0.07, 0.17], $t(171.33) = 0.82$, $p = .415$. Topic familiarity was positively associated with trustworthiness, $\beta = 0.87$, 95% CI [0.79, 0.94], $t(1660.01) = 22.92$, $p < .001$. Without the covariate, the effect of debiasing on headline trustworthiness remained non-significant, $\beta = 0.06$, 95% CI [-0.07, 0.19], $t(1558.27) = 0.90$, $p = .368$.

***Comprehensiveness.*** There was no significant effect of debiasing on perceived comprehensiveness (whether the headline tells the whole story), $\beta = -0.01$, 95% CI [-0.13, 0.10], $t(168.39) = -0.23$, $p = .817$. Topic familiarity was positively associated with comprehensiveness, $\beta = 0.69$, 95% CI [0.62, 0.77], $t(1542.49) = 18.44$, $p < .001$. Without the covariate, the effect of debiasing on comprehensiveness remained non-significant, $\beta = -0.01$, 95% CI [-0.13, 0.12], $t(1371.84) = -0.09$, $p = .931$.

***Openness to Consider Perspective.*** Debiasing did not significantly predict openness to consider the article's perspective, β = 0.04, 95% CI [-0.08, 0.16], $t(167.90) = 0.72$, $p = .474$. Topic familiarity was positively associated with openness to consider the article's perspective, β = 0.83, 95% CI [0.75, 0.90], $t(1578.66) = 21.18$, $p < .001$. Without the covariate, the effect of debiasing on openness to consider the article's perspective remained non-significant, β = 0.05, 95% CI [-0.08, 0.19], $t(169.14) = 0.78$, $p = .437$.

All pre-registered condition effects were robust to rated semantic equivalence: re-estimating each model with the pair's mean equivalence rating added as a grand-mean-centred article-level covariate left every condition estimate essentially unchanged (e.g., perceived anti-conservative bias β = −0.09, $p = .167$ at baseline vs. β = −0.092, $p = .165$ with the covariate; full comparison in Supplementary Table S1.7).

**Exploratory Tests: Moderation Analyses.** We next examined whether individual differences moderated the effect of debiasing by simultaneously adding three two-way interactions (condition × media trust, condition × cognitive flexibility, condition × in-group identification) to the models, while continuing to control for topic familiarity.

***Anti-Conservative Bias (Manipulation Check).*** None of the interactions between condition and individual difference measures were significant (all $p$s > .15).

***Headline Trustworthiness.*** None of the interactions between condition and individual difference measures were significant (all $p$s > .09).

***Comprehensiveness.*** None of the interactions between condition and individual difference measures were significant (all $p$s > .18).

***Openness to Consider Perspective.*** The interaction between condition and cognitive flexibility was significant, β = -0.14, 95% CI [-0.26, -0.02], $t(161.96) = -2.34$, $p = .021$, indicating that the effect of debiasing on openness varied by cognitive flexibility. The interactions with media trust ($p = .242$) and in-group identification ($p = .198$) were not significant in the same model. To probe the statistically significant interaction, we conducted a Johnson-Neyman analysis to identify the range of cognitive flexibility scores at which the effect of debiasing on openness to consider the article's perspective was statistically significant. The Johnson-Neyman analysis (Figure 2) revealed that the effect of condition became significant ($p < .05$) for cognitive flexibility scores below 4.83. Given that the observed range of cognitive flexibility scores was 3.00 to 7.00, this threshold falls within the observed range of the moderator, indicating that the interaction was practically meaningful. Simple slopes analysis confirmed that the effect of debiasing on openness to consider the article's perspective was significant at low levels of cognitive flexibility (−1 *SD*: β = 0.19, $p = .032$), but not at mean (β = 0.04, $p = .471$) or high (+1 *SD*: β = −0.10, $p = .257$) levels, with the effect of debiasing weakening and descriptively reversing as cognitive flexibility increased.

**Figure 2**

*Johnson-Neyman Analysis for the Moderating Effect of Cognitive Flexibility on the Effect of Debiasing Condition on Openness to Consider Perspective in the Human Sample*

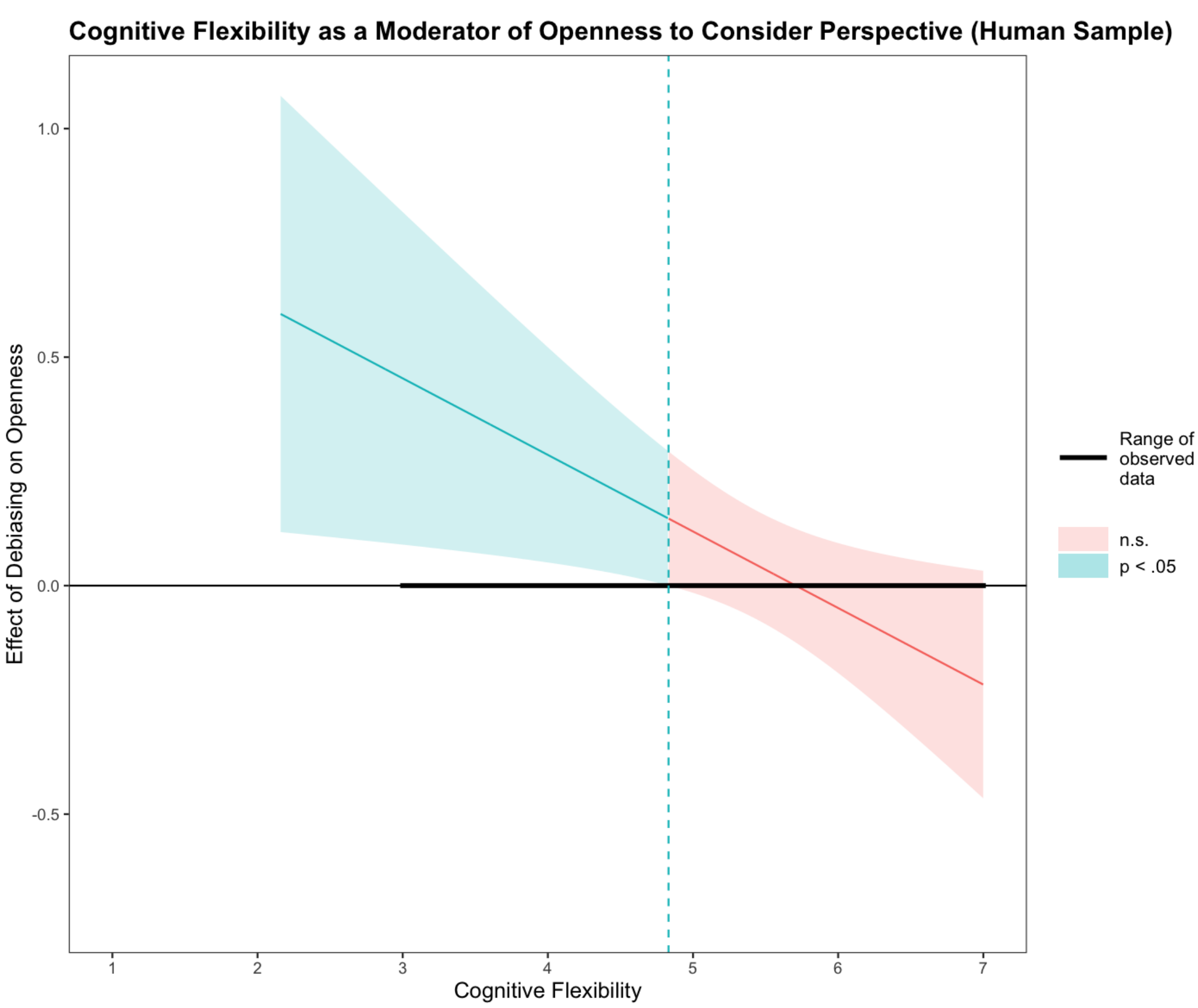

*Note.* The solid line represents the conditional effect of debiasing at each level of cognitive flexibility. Shaded regions indicate 95% confidence bands. The dashed vertical line marks the Johnson-Neyman threshold, below which the conditional effect is significant at $p < .05$. The threshold was 4.83. The black horizontal line indicates the range of observed cognitive flexibility scores (3.00 to 7.00).

### *Silicon Sample Results*

We conducted the same mixed-effects linear regression models as for human participants (see Figure 3).

**Figure 3**

*Mean Ratings by Silicon Participants for Original and Debiased Articles for the Manipulation Check and Three Dependent Variables Measured*

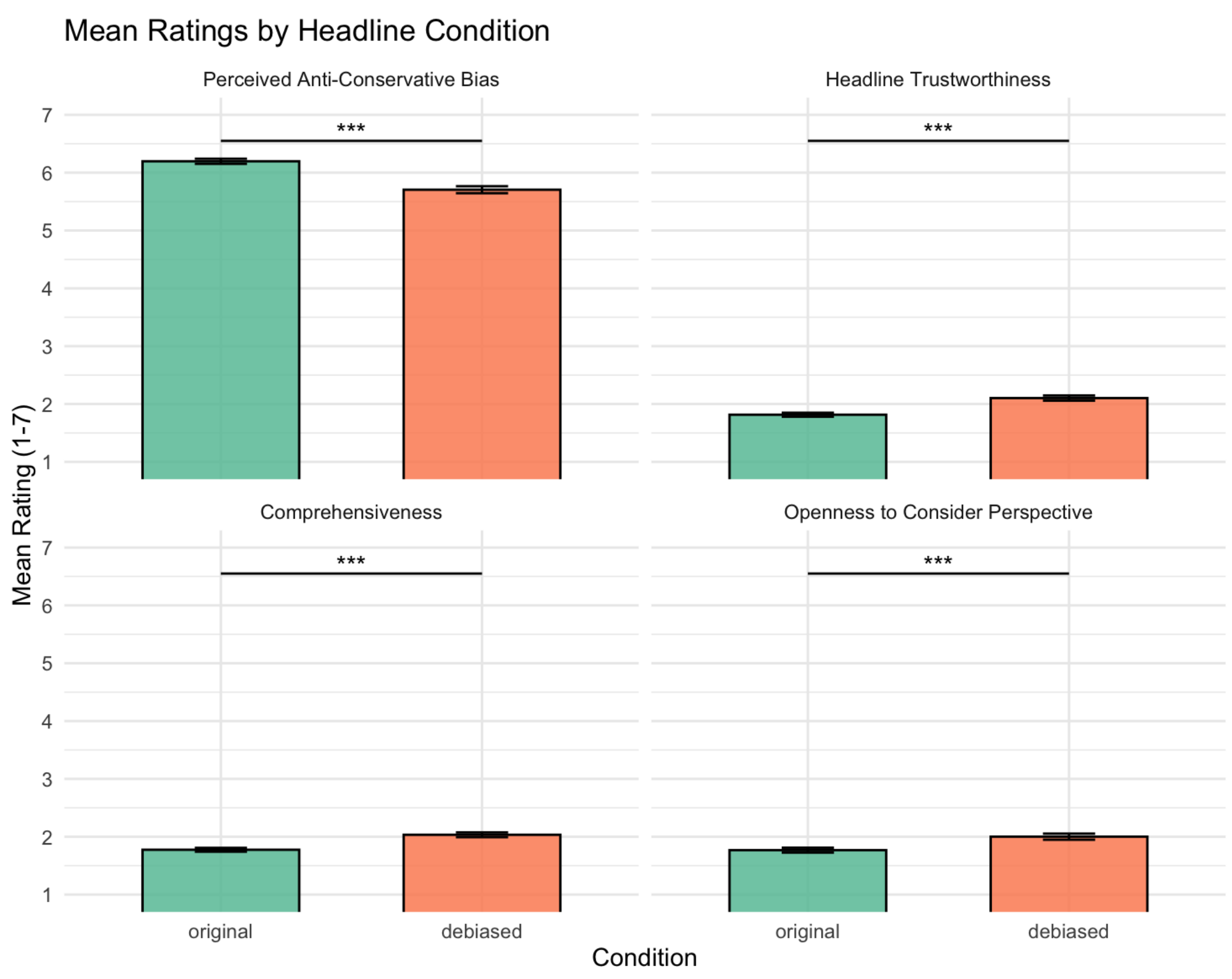

*Note.* Error bars represent 95% CI. ***$p < .001$. Anti-conservative bias is measured on a scale from 0 to 6 and was transformed to 1 to 7 prior to visualization.

**Confirmatory Tests.** We estimated the same pre-registered models for the silicon sample.

***Anti-Conservative Bias (Manipulation Check).*** In contrast to the human sample, the debiasing manipulation significantly reduced perceived anti-conservative bias among silicon participants, with a sizeable effect, β = -0.44, 95% CI [-0.49, -0.39], $t(1063.27) = -16.84$, $p < .001$.Debiased articles were perceived as less biased than original articles. Topic familiarity was not significantly associated with perceived bias, β = 0.01, 95% CI [-0.02, 0.04], $t(1420.07) = 0.60$, $p = .551$. Without the familiarity covariate, the effect of debiasing on perceived bias remained significant, β = -0.44, 95% CI [-0.49, -0.39], $t(1038.64) = -16.80$, $p < .001$.

***Headline Trustworthiness.*** The debiasing condition significantly increased perceived trustworthiness, β = 0.27, 95% CI [0.23, 0.32], $t(1448.28) = 12.70$, $p < .001$, with debiased articles rated as more trustworthy than original articles. Topic familiarity was positively associated with trustworthiness, β = 0.03, 95% CI [0.01, 0.06], $t(1669.88) = 2.55$, $p = .011$. Without the covariate, the effect of debiasing on trustworthiness remained significant, β = 0.27, 95% CI [0.23, 0.32], $t(1457.92) = 12.69$, $p < .001$.

***Comprehensiveness.*** There was a significant effect of debiasing on perceived comprehensiveness, β = 0.25, 95% CI [0.21, 0.29], $t(1630.26) = 12.61$, $p < .001$, with debiased articles rated as more complete than original articles. Topic familiarity was positively associated with comprehensiveness, β = 0.03, 95% CI [0.00, 0.05], $t(1689.93) = 1.99$, $p = .047$. Without the covariate, the effect of debiasing remained significant, β = 0.25, 95% CI [0.21, 0.29], $t(174.15) = 12.57$, $p < .001$.

***Openness to Consider Perspective.*** Debiasing significantly predicted openness to consider the article's perspective, β = 0.29, 95% CI [0.25, 0.34], $t(974.91) = 12.19$, $p < .001$, with silicon participants showing greater openness to consider the article's perspective toward debiased articles. Topic familiarity was not significantly associated with openness to consider the article's perspective, β = 0.02, 95% CI [-0.01, 0.05], $t(1733.82) = 1.23$, $p = .218$. Without the covariate, the effect of debiasing remained significant, β = 0.29, 95% CI [0.25, 0.34], $t(982.94) = 12.19$, $p < .001$.

To test whether the silicon findings are specific to o3-mini, we re-ran the full silicon pipeline with five additional models spanning scale and architecture (GPT-4o mini, GPT-4o, o3, GPT-5, and GPT-5 mini). All four pre-registered effects replicated at $p < .001$ in every model — in sharp contrast to the uniformly null human effects (Figure 4; Supplementary Section S3). Debiasing effects on perceived bias, for example, ranged from −0.24 to −0.44 across models while the human effect was −0.10 ($p = .20$). Note that all per-sample estimates in this paragraph are simple effects from a single combined robustness model fitted jointly to the human and all six silicon samples; they therefore differ slightly from the pre-registered human-only models reported above (which give β = −0.09, $p = .167$ for perceived bias).

**Figure 4**

*Silicon-model robustness, Study 1: debiasing effect (debiased minus original) on each outcome for the human sample and each of the six LLMs, with 95% confidence intervals.*

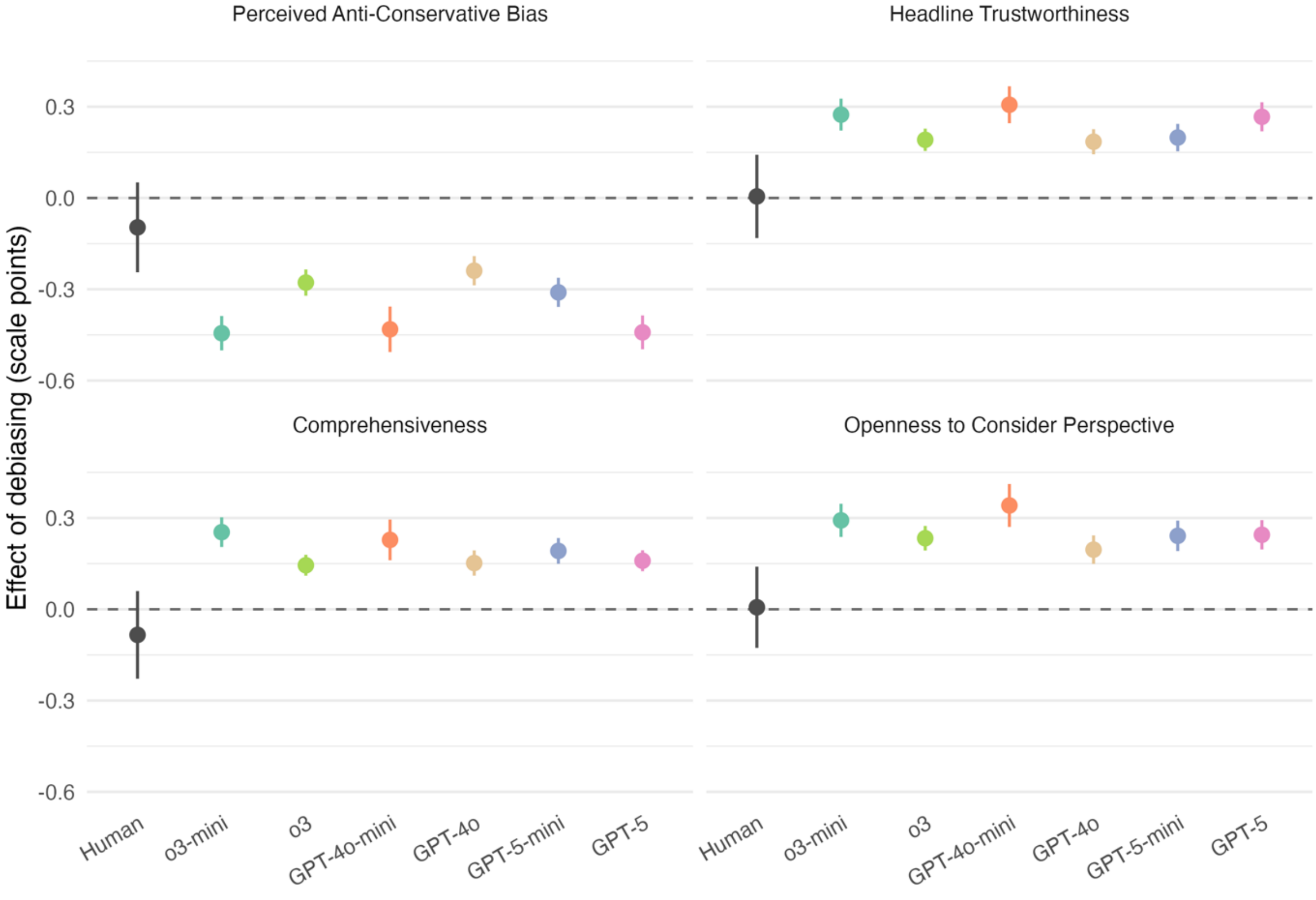

**Exploratory Tests: Moderation Analyses.** To explore whether individual differences moderated the effect of debiasing, we estimated the exact same models as for human participants. Across the four outcomes, only one interaction reached the significance threshold: the interaction between condition and media trust on headline trustworthiness ($p = .050$).

***Anti-Conservative Bias (Manipulation Check).*** None of the interactions between condition and individual difference measures were significant (all $ps \geq .298$).

***Headline Trustworthiness.*** The interaction between condition and media trust was significant at the threshold, $\beta = 0.04$, 95% CI [0.00, 0.09], $t(1640.36) = 1.96$, $p = .050$, indicating that the effect of debiasing on trustworthiness varied modestly by media trust. The interactions with cognitive flexibility, $\beta = 0.02$, 95% CI [-0.02, 0.07], $t(1634.35) = 1.09$, $p = .277$, and in-group identification, $\beta = -0.01$, 95% CI [-0.05, 0.04], $t(1637.87) = -0.29$, $p = .773$, were not significant.

To probe the significant interaction, we conducted a Johnson-Neyman analysis (Figure 5) to identify the range of media trust scores at which the effect of debiasing on trustworthiness was statistically significant. The analysis revealed that the effect of condition became significant ($p < .05$) for media trust scores above −1.08. Given that the observed range of media trust scores was 1.00 to 7.00, this threshold falls below the observed range of the moderator, indicating that the debiasing effect on trustworthiness was significant across the entire observed range of media trust. Simple slopes analysis confirmed that the effect of debiasing on trustworthiness was significant at low (−1 *SD*: $\beta = 0.23$, $p < .001$), mean ($\beta = 0.27$, $p < .001$), and high (+1 *SD*: $\beta = 0.32$, $p < .001$) levels of media trust, with the effect of debiasing becoming stronger as media trust increased.

***Comprehensiveness.*** None of the interactions between condition and individual difference measures were significant (media trust: $\beta = 0.03$, 95% CI [−0.01, 0.07], $t(1452.08)$ all $ps \geq .181$).

**Figure 5**

*Johnson-Neyman Analysis for the Moderating Effect of Media Trust on the Effect of Debiasing Condition on Trustworthiness) in the o3 mini Silicon Sample*

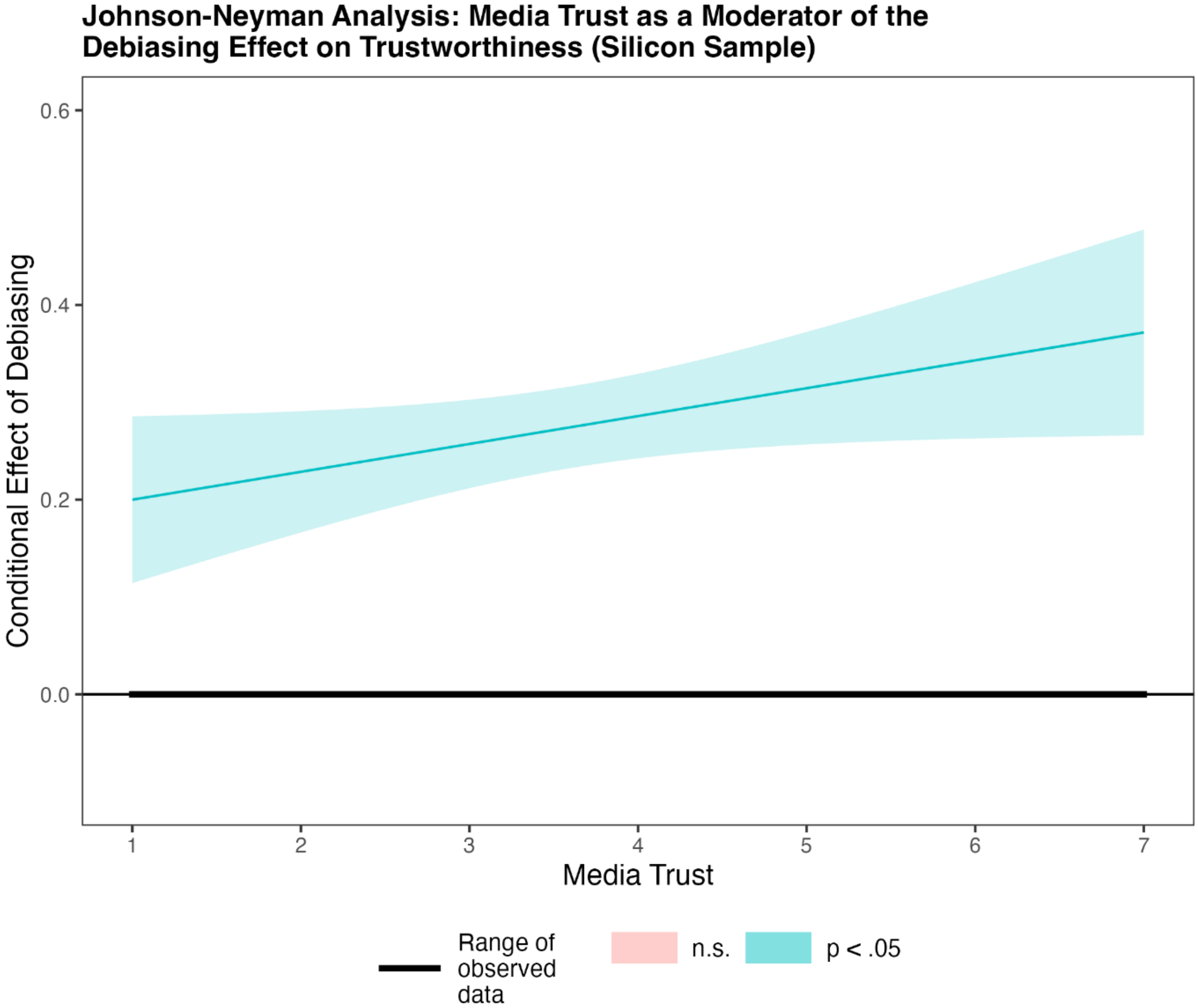

*Note.* The solid line represents the conditional effect of debiasing at each level of media trust. Shaded regions indicate 95% confidence bands. The dashed vertical line marks the Johnson-Neyman threshold, above which the conditional effect is significant at $p < .05$. The threshold was −1.08; as this falls below the observed minimum of media trust scores (1.00), the effect of debiasing was significant across the entire observed range (1.00 to 7.00).

***Openness to Consider Perspective.*** None of the interactions between condition and individual difference measures were significant (all $p$s $\geq .151$).

### *Direct Comparison Between Human and Silicon Participants*

To directly test whether the effect of debiasing differed between human and silicon participants, we combined both samples and fitted mixed-effects models with condition (original vs. debiased), sample type (human vs. silicon), and their interaction as predictors, with random intercepts and slopes for condition by participant and random intercepts by trial. Models were estimated both with and without topic familiarity as a covariate to gauge their robustness.

**Anti-Conservative Bias (Manipulation Check).** The interaction between condition and sample type was significant, β = -0.36, 95% CI [-0.50, -0.22], $t(326.09) = -5.03$, $p < .001$, confirming that the effect of debiasing on perceived bias was significantly larger for silicon participants than for human participants. Without the familiarity covariate, the interaction remained significant, β = -0.36, 95% CI [-0.50, -0.22], $t(326.53) = -5.02$, $p < .001$

**Headline Trustworthiness.** The interaction between condition and sample type was significant, β = 0.22, 95% CI [0.09, 0.35], $t(344.33) = 3.30$, $p = .001$, indicating that the positive effect of debiasing on trustworthiness was significantly larger for silicon participants than for human participants. Without the familiarity covariate, the interaction remained significant, β = 0.22, 95% CI [0.09, 0.36], $t(344.93) = 3.19$, $p = .002$.

**Comprehensiveness.** The interaction between condition and sample type was significant, β = 0.26, 95% CI [0.14, 0.39], $t(338.31) = 4.11$, $p < .001$, indicating that the effect of debiasing on perceived comprehensiveness was significantly larger for silicon participants than for human participants. Without the familiarity covariate, the interaction remained significant, β = 0.26, 95% CI [0.13, 0.39], $t(2721.41) = 3.99$, $p < .001$.

**Openness to Consider Perspective.** The interaction between condition and sample type was significant, β = 0.25, 95% CI [0.11, 0.39], $t(338.35) = 3.60$, $p < .001$, confirming that the effect of debiasing on openness to consider the article's perspective was significantly larger for silicon participants than for human participants. Without the familiarity covariate, the interaction remained significant, β = 0.25, 95% CI [0.11, 0.39], $t(338.65) = 3.41$, $p = .001$.

In summary, across all four outcomes, the condition × sample interactions were significant, demonstrating that the debiasing manipulation had a significantly larger effect on silicon participants than on human participants. These results were robust to the inclusion of topic familiarity as a covariate.

**Preliminary Discussion**

The present study examined whether AI-generated debiasing of left-leaning news articles could reduce perceived bias and improve receptiveness among conservative readers. Our findings reveal a striking disconnect between human- and AI simulation-based debiasing effectiveness at the level of subtle linguistic substitutions implemented here.

Contrary to our pre-registered hypotheses, the debiasing manipulation had no statistically significant effect on human participants' perceptions of bias, trustworthiness, comprehensiveness, or openness to consider the article's perspective. Conservative readers did not differentiate between original MSNBC opinion articles and versions that had been linguistically "softened" by o3-mini through word substitutions designed to appeal to a conservative audience. This null result is notable given prior evidence that neutralizing loaded language can increase perceived fairness among opposing partisans (Walker et al., 2025). However, those interventions typically involved more substantive restructuring of content than the minimal lexical substitutions tested here. Our findings suggest that the kind of surface-level word replacement by LLMs may be insufficient to overcome the defensive processing triggered when partisans encounter out-group media sources.

Strikingly, silicon participants (o3-mini models prompted with exactly matched human demographic and attitudinal profiles but blind to the experiment just as humans were) were responsive to the debiasing manipulation across all outcomes. Here, debiased articles were perceived as significantly less biased, more trustworthy, more comprehensive, and more deserving of consideration. Direct statistical comparison confirmed that these differences were not merely descriptive: the condition × sample interactions were significant for all four outcomes, indicating that the debiasing effect was reliably larger for silicon participants than for humans. This pattern aligns with the broader concern that LLMs rely on statistical regularities that may diverge from the causal mechanisms governing human cognition (Bender et al., 2021; Mitchell & Krakauer, 2023), and with recent findings that LLM predictions of intervention efficacy tend to overestimate effect sizes (Cui et al., 2025).

Moderation analyses further underscored this asymmetry. In the human sample, where the debiasing manipulation produced no main effects, the only significant moderation was a condition × cognitive flexibility interaction on openness ($p = .021$), where debiasing increased openness only among participants with lower cognitive flexibility. No other individual-difference interactions reached significance across the four outcomes. While a null main effect can sometimes mask heterogeneity, such as when a variable produces effects in opposing directions across subgroups, a single isolated interaction across twelve tests is difficult to distinguish from chance and does not alter the broader picture: the manipulation was largely ineffective, cutting across the individual-difference variables we measured rather than being confined to a particular subgroup.

In the silicon sample, where the debiasing manipulation produced robust main effects, moderation was likewise largely absent. Only the condition × media trust interaction on headline trustworthiness reached the significance threshold ($\beta = 0.04$, $p = .050$), with the debiasing effect strengthening modestly as media trust increased and remaining significant across the entire observed range of media trust scores. No other interactions reached significance across the four outcomes. The human–silicon misalignment in Study 1 is therefore carried almost entirely by the main effects: the silicon sample responded strongly and uniformly to a manipulation that left human judgments unchanged.

Overall, our results in Study 1 suggest that LLMs may operate with an implicit theory of bias that does not align with how humans actually experience partisan content. The debiasing strategy (replacing emotionally charged words with more moderate synonyms) reflects a surface-level, lexical understanding of bias. The model seems to overestimate its effectiveness both in terms of the importance of word choice in shaping perceived bias and human sensitivity to such linguistic cues. This asymmetry extends beyond the debiasing manipulation itself. In the human sample, topic familiarity was the strongest predictor of headline trustworthiness ($\beta = 0.87$), yet in the silicon sample, the same association was negligible ($\beta = 0.03$). The model thus lacks an adequate representation of how prior exposure to an issue anchors credibility judgments, further illustrating the gap between its implicit psychology and the mechanisms that govern human information processing. For human readers, perceived bias may be driven more by deeper features such as the ideological framing of the argument. If perceived bias is primarily a top-down projection of identity (Mohanty et al., 2025) rather than a stimulus-driven reaction to specific textual cues (Caparos et al., 2015), then interventions that leave the underlying argumentative frame intact should fail regardless of how thoroughly individual words are softened. This interpretation is consistent with our human null findings and motivates the more substantive reframing intervention tested in Study 2, which targets not merely word choice but the framing of the issue itself.

## Study 2

Study 1 tested whether minimal lexical changes to partisan news content could shift conservative readers' trust-relevant judgments. Study 2 extended this investigation in two ways. First, given the lack of statistically significant effects in the human sample in the first study, we tested a more substantive debiasing intervention. Whereas Study 1 employed targeted word-level replacements that preserved the original narrative structure, the intervention in Study 2 involved broader reframing of the headlines, altering not just emotive language but the way the issue itself was linguistically presented. If minimal debiasing proved insufficient in Study 1, this more invasive intervention would test whether deeper structural changes could overcome defensive processing among out-group readers.

Second, and critically, Study 2 included both conservative and liberal participants. This allowed us to examine a question that the first study could not address: whether debiasing effects are symmetrical across the political spectrum, or whether a backfire effect occurs among the ideologically aligned in-group. As discussed in the introduction, partisan media consumers often seek validation rather than neutrality, valuing content that defends their worldview and signals shared moral commitments (Finkel et al., 2020). Consequently, an MSNBC headline stripped to some extent of its liberal framing may be perceived by liberal readers not as balanced but as diluted or as engaging in false equivalence. By including liberal participants, Study 2 tested whether the optimality of a debiased text is relative, such that a successful intervention for one group constitutes a degradation for the other.

As in Study 1, we predicted that debiased headlines would elicit higher ratings of trustworthiness, perceptions of telling the whole story, and willingness to consider the presented perspective among conservative participants. We further explored whether the effect of debiasing would differ in valence among liberal (as compared to conservative) participants, and whether effects would be moderated by baseline media trust, cognitive flexibility, and strength of partisan identification. The silicon participant procedure was again run in parallel to assess LLM-human alignment under the expanded design.

## Method

### *Human Participants*

Based on a Monte Carlo power simulation, 170 participants per group (i.e., 340 total) were needed to provide a 90% chance to observe a three-way cross-level interaction (debiasing manipulation × political group × continuous moderator) for an effect size of Cohen's $d = 0.25$ at a significance criterion of 0.05. Therefore, 186 Republican and 181 Democrat US participants were recruited from Prolific with the same quality requirement as in Study 1. The experiment took 12 minutes on average.

Participants indicated that they were from the U.S., above 18 years of age and spoke English fluently. Of the 367 participants who completed the study, 49.7% identified as women and 50.3% as men. In terms of education, 7.8% had completed high school or equivalent, 14.7% had some college experience without obtaining a degree, 10.3% held an associate or technical degree, 39.1% held a bachelor's degree, and 28.2% held a graduate or professional degree (e.g., MA, PhD, JD, MD). Regarding racial identification, 74.7% identified as White, 18.7% as Black or African American, 2.6% as Asian, 1.1% as American Indian or Alaska Native, and 1.1% selected Other; the remaining 1.8% identified with multiple racial categories. Some participants selected multiple racial categories, including 0.6% who identified as both White and American Indian/Alaska Native, 0.3% as both White and Asian, 0.6% as both White and Black or African American, and 0.3% as White and Other. The mean age was 45.13 years ($SD = 14.03$, range: 19–88). Study 2 was deployed on Prolific on 28 July 2025. The bulk of responses (345 of 348) were collected the same day, with the remaining 3 Democrat responses submitted in the early hours of 29 July 2025 (UTC). We note that the articles used in Study 2 were the same set originally scraped on 26 April 2025 for Study 1, introducing a roughly three-month lag between scraping and data collection. However, given the volume and pace of news cycles in contemporary US media, it is unlikely that participants retained detailed memory of the original framing of specific opinion pieces from three months prior. To further address the possibility that prior exposure or issue familiarity could still influence evaluations, topic familiarity was included as a pre-registered control variable in all analyses (as in Study 1).

Nineteen participants failed two or more of three attention checks. Exclusions based on this pre-registered requirement left us with 176 conservative and 172 liberal participants ($N$ = 348) who completed 3,480 trials in total. Sixteen of these participants had incomplete moderator data; analyses involving moderators (Stages 3 and 4) were therefore conducted with the remaining 332 participants (3,320 trials). A further 17 bias trials from two participants were missing due to non-numeric responses; models predicting anti-conservative bias were therefore fit to 3,463 trials in Stages 1–2 and 3,303 trials in Stages 3–4.

***Silicon Participants***

The silicon participant procedure was identical to Study 1, with the sample exactly matching the human sample. A total of 332 simulated participants were created, matching the human participants for whom complete moderator data were available, maintaining a one-to-one mapping. The procedure yielded 3,320 valid trials.

***Procedure***

The same ten MSNBC opinion headlines from Study 1 served as base stimuli. We weighed using these original stimuli to maintain comparability against introducing new stimuli to enhance timeliness, ultimately opting for the former since data collection remained close to the publication dates of the headlines. The key procedural difference from Study 1 concerned the debiasing intervention. Whereas Study 1 instructed o3-mini to replace emotive words with moderate synonyms while preserving narrative structure, Study 2 instructed the model to reframe the headline so that the underlying issue was presented in a way that would be accessible to and be perceived as fair by a conservative audience. This allowed the model to alter not only word choice but the framing of the issue itself. Specifically, Study 2's prompt asked the model to "rephrase the headline and description in a way that lowers emotional intensity and aligns with a conservative communication style," explicitly permitting sentence-level rephrasing and passive constructions; its worked examples model the resulting attribution style ("seen as inaccurate…by critics," "concerns grow over"), which in the stimuli surfaces as hedged, attributed characterizations ("viewed as permitting discrimination"). For the full prompts and the differences between original and reframed headlines, see Table 1 above, and the supplementary materials. The trial structure, dependent variables, moderators, attention checks, and manipulation check were otherwise identical to Study 1, with one adaptation: the in-group identification items and manipulation check were reframed to match participants' political group (referencing "conservative people" for Republicans and "liberal people" for Democrats). Cronbach's alphas were as follows: media trust ($\alpha = .98$), cognitive flexibility ($\alpha = .76$), and in-group identification ($\alpha = .92$).

As in Study 1, an independent sample of Republican/conservative raters ($N = 88$) evaluated the Study 2 headline pairs side by side. Semantic equivalence was again rated toward the identical-meaning pole ($M = 5.31$, $SD = 1.52$), and 38.1% of judgments indicated a factual-claim change — a higher rate than Study 1's, consistent with the deliberately deeper reframing intervention. For the Study 2 stimuli, the manipulation check embedded in the main experiment confirmed the successful manipulation: debiased headlines were rated as significantly less biased ($\beta = -0.16$, $p = .003$; among conservative participants, $\beta = -0.34$, $p < .001$). The validation raters' single-headline ecological validity judgments likewise showed no statistically significant original/debiased difference ($M = 4.55$ vs. 4.62, $p = .206$).

***Data Analysis***

The analytic strategy followed Study 1, with extensions to accommodate the two-group design. Analyses proceeded in four stages, as pre-registered. First, we estimated models testing the main effect of condition (original vs. reframed) across all participants. Second, we added political group (conservative vs. liberal) and its interaction with condition to test whether the effect of debiasing differed between groups. Third, we tested two-way interactions between condition and each of the three continuous moderators (media trust, cognitive flexibility, and in-group identification). Fourth, we tested whether these two-way interactions were further qualified by political group, yielding three-way condition × moderator × political group interactions. Separate models were estimated for each dependent variable. All other analytic details (software, random effects structure, covariates, assumption testing, and bootstrapping procedure) were as described in Study 1.

We also ran the same models with silicon participants generated above. We additionally pooled the human and silicon datasets to directly test whether the effect of debiasing differed between samples. These models included a condition × sample interaction to test for overall differences in effect magnitude, a condition × sample × political group interaction to test whether any human-silicon differences varied between conservatives and liberals, and a condition × sample × media trust interaction to test whether the moderating role of media trust differed across samples. Coefficients are reported as in Study 1. Simple effects and simple slopes reported with $z$-values were obtained from emmeans-based contrasts.

All data, materials, and code can be obtained at https://osf.io/c2fpe/overview?view_only=02561ad5eb08446d877760a9aad35d88 (pre-

registration), and https://osf.io/na78b/overview?view_only=8503c0c0d91649db80f903e05d5ba5b3 (code, materials).

## Results

### *Human Sample Results*

As in Study 1, we conducted mixed-effects linear regression models predicting each outcome from condition (original vs. reframed) and topic familiarity as a covariate, with random intercepts and slopes for condition by participant and random intercepts by trial. Results are presented following the four pre-registered stages. Figure 6 displays marginal means by condition, split by political group for outcomes where the Condition × Political Group interaction was significant (perceived bias and trustworthiness) and pooled across participants for outcomes where it was not (comprehensiveness and openness).

**Figure 6**

*Marginal Means by Condition for the Manipulation Check and Three Dependent Variables Measured in Study 2, Split by Political Group Where Interactions Were Significant*

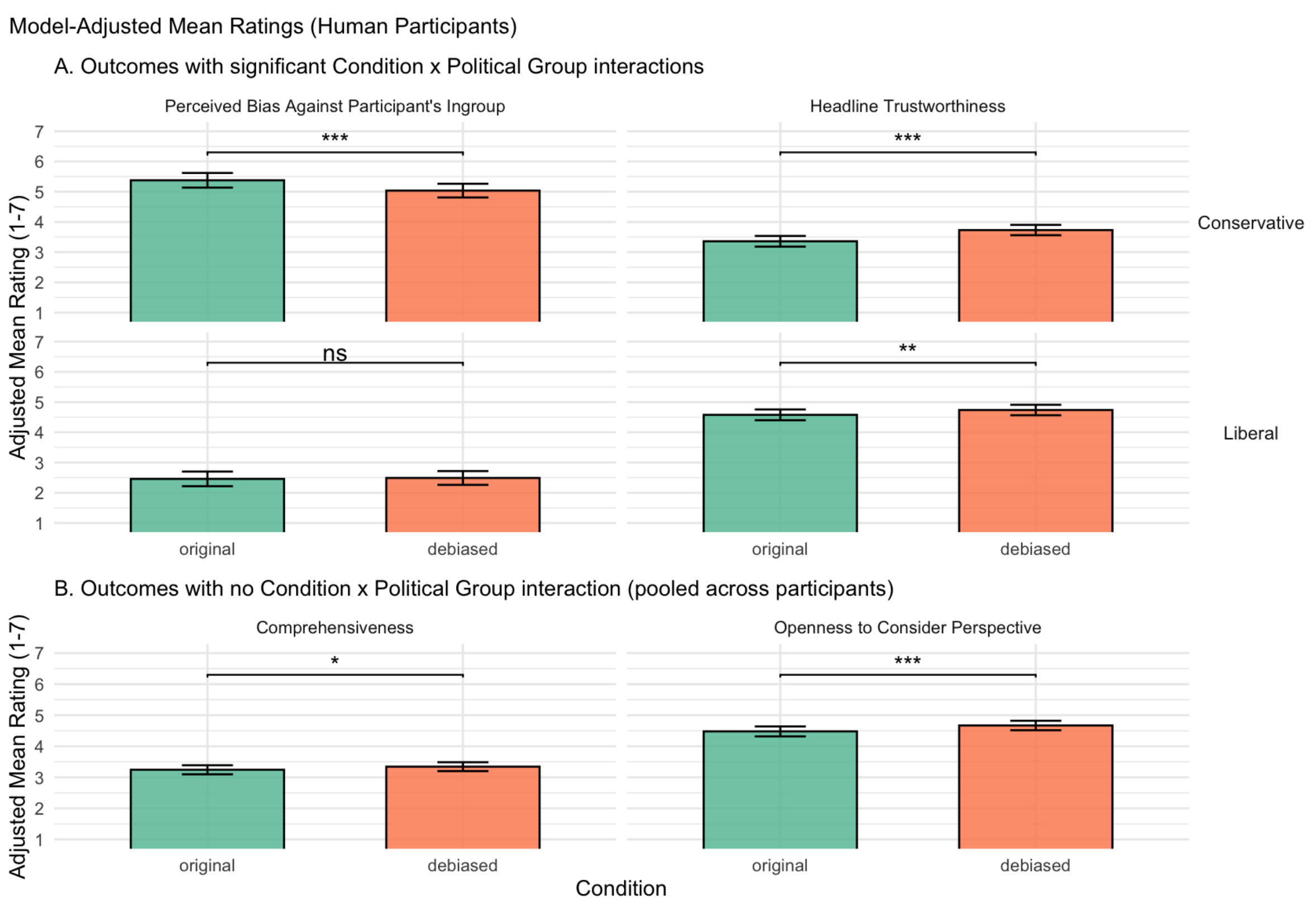

*Note.* Bars display estimated marginal means from mixed-effects models with topic familiarity held at its sample mean. Panel A shows outcomes for which the Condition × Political Group interaction was significant (perceived bias against participants' ingroup and headline trustworthiness), with results split by political group. Panel B shows outcomes for which the interaction was not significant (comprehensiveness and openness to consider perspective), with results pooled across all participants. Perceived bias against participants' ingroup was originally measured on a 0–6 scale and rescaled to 1–7 for comparability with the other outcomes. Error bars represent 95% CIs. ***$p < .001$, **$p < .01$, *$p < .05$, ns = not significant.

**Confirmatory Tests.** We first report the pre-registered main-effect and political-group models (Stages 1 and 2).

**Main Effects (Stage 1).** Across all participants, the reframing manipulation significantly affected all four outcomes.

**Anti-Conservative Bias (Manipulation Check).** Debiased headlines were perceived as significantly less biased, β = −0.16, 95% CI [−0.26, −0.06], $t(341.36) = -3.01$, $p = .003$. Topic familiarity was not significantly associated with perceived bias (β = −0.04, 95% CI [−0.09, 0.02], $t(3086.37) = -1.32$, $p = .189$).

**Headline Trustworthiness.** Debiasing significantly increased perceived trustworthiness, β = 0.27, 95% CI [0.19, 0.35], $t(341.97) = 6.63$, $p < .001$. Topic familiarity was positively associated with trustworthiness (β = 0.89, 95% CI [0.84, 0.94], $t(3100.02) = 35.76$, $p < .001$).

**Comprehensiveness.** There was a significant effect of debiasing on perceived comprehensiveness, β = 0.10, 95% CI [0.02, 0.18], $t(2890.88) = 2.45$, $p = .014$. Topic familiarity was positively associated with comprehensiveness (β = 0.71, 95% CI [0.66, 0.76], $t(2574.87) = 27.43$, $p < .001$).

**Openness to Consider Perspective.** Debiasing significantly predicted greater openness to consider the article's perspective, β = 0.19, 95% CI [0.11, 0.28], $t(2162.48) = 4.41$, $p < .001$. Topic familiarity was positively associated with openness to consider the article's perspective (β = 0.62, 95% CI [0.57, 0.68], $t(2846.66) = 22.80$, $p < .001$).

The Study 2 condition effects likewise held with the semantic-equivalence covariate included (e.g., headline trustworthiness β = 0.27, $p < .001$ in both specifications; perceived bias β = −0.16, $p = .003$ in both; Supplementary Table S1.7).

**Political Group Moderation (Stage 2).** The condition × political group interaction was significant for perceived anti-conservative bias, β = 0.37, 95% CI [0.17, 0.57], $t(342.16) = 3.59$, $p < .001$, and headline trustworthiness, β = −0.21, 95% CI [−0.37, −0.05], $t(345.15) = -2.63$, $p = .009$, but not for openness to consider perspective, β = −0.17, 95% CI [−0.34, 0.003], $t(2414.49) = -1.93$, $p = .054$, or comprehensiveness, β = −0.10, 95% CI [−0.26, 0.06], $t(2990.76) = -1.20$, $p = .230$.

Simple effects analyses revealed that conservative participants drove the bulk of the pooled effects. Among conservatives, debiased headlines were perceived as significantly less biased (β = −0.34, 95% CI [−0.48, −0.20], $z = 4.67$, $p < .001$), and more trustworthy (β = 0.37, 95% CI [0.26, 0.48], $z = 6.57$, $p < .001$), and comprehensive (β = 0.15, 95% CI [0.04, 0.26], $z = 2.58$, $p = .010$). Among liberal participants, the effect of debiasing was nonsignificant for perceived bias (β = 0.03, 95% CI [−0.11, 0.17], $z = 0.42$, $p = .677$) and weaker yet still significant for headline trustworthiness (β = 0.16, 95% CI [0.05, 0.27], $z = 2.83$, $p = .005$).

**Exploratory Tests: Moderation Analyses.** We next examined moderation by individual differences (Stages 3 and 4).

**Moderation by Individual Differences (Stage 3).** Media trust significantly moderated the effect of debiasing on perceived bias against participants' political ingroup, β = 0.14, 95% CI [0.03, 0.24], $t$(325.02) = 2.48, $p$ = .014, headline trustworthiness, β = −0.098, 95% CI [−0.18, −0.02], $t$(322.23) = −2.32, $p$ = .021, and openness to consider perspective, β = −0.09, 95% CI [−0.18, −0.002], $t$(2028.04) = −2.00, $p$ = .046, but not comprehensiveness, β = −0.025, 95% CI [−0.110, 0.060], $t$(2725.67) = −0.58, $p$ = .564. Although media trust statistically moderated the effect of debiasing on perceived bias against participants' ingroup, headline trustworthiness, and openness to consider perspective, follow-up Johnson-Neyman analyses indicated that the substantive reach of these moderations differed across outcomes (see Figure 7). For perceived bias, the conditional effect of debiasing was significant for media trust scores below 4.05. Simple slopes confirmed that debiasing significantly reduced perceived bias at low (−1 *SD*: β = −0.30, $p$ < .001) and mean (β = −0.17, $p$ = .002) levels of media trust, but not at high (+1 *SD*: β = −0.03, $p$ = .698) levels. For headline trustworthiness, the conditional effect was significant below 5.59, and simple slopes indicated that debiasing increased trustworthiness at low (−1 *SD*: β = 0.36, $p$ < .001), mean (β = 0.26, $p$ < .001), and high (+1 *SD*: β = 0.17, $p$ = .005) levels of media trust, though the effect attenuated as media trust increased. For openness to consider perspective, the conditional effect was significant below 4.93. Simple slopes confirmed that debiasing increased openness at low (−1 *SD*: β = 0.29, $p$ < .001) and mean (β = 0.20, $p$ < .001) levels of media trust, but not at high (+1 *SD*: β = 0.11, $p$ = .080) levels. Across all three outcomes, the pattern was consistent: the effect of debiasing was strongest among participants with lower media trust and attenuated or disappeared as media trust increased. Neither cognitive flexibility nor in-group identification moderated the effect of debiasing on any outcome (all $p$s ≥ .387).

**Figure 7**

*Johnson-Neyman Analyses for the Moderating Effect of Media Trust on the Effect of Debiasing Condition on Perceived Bias Against Participants' Ingroup, Headline Trustworthiness, and Openness to Consider Perspective in the Human Sample of Study 2*

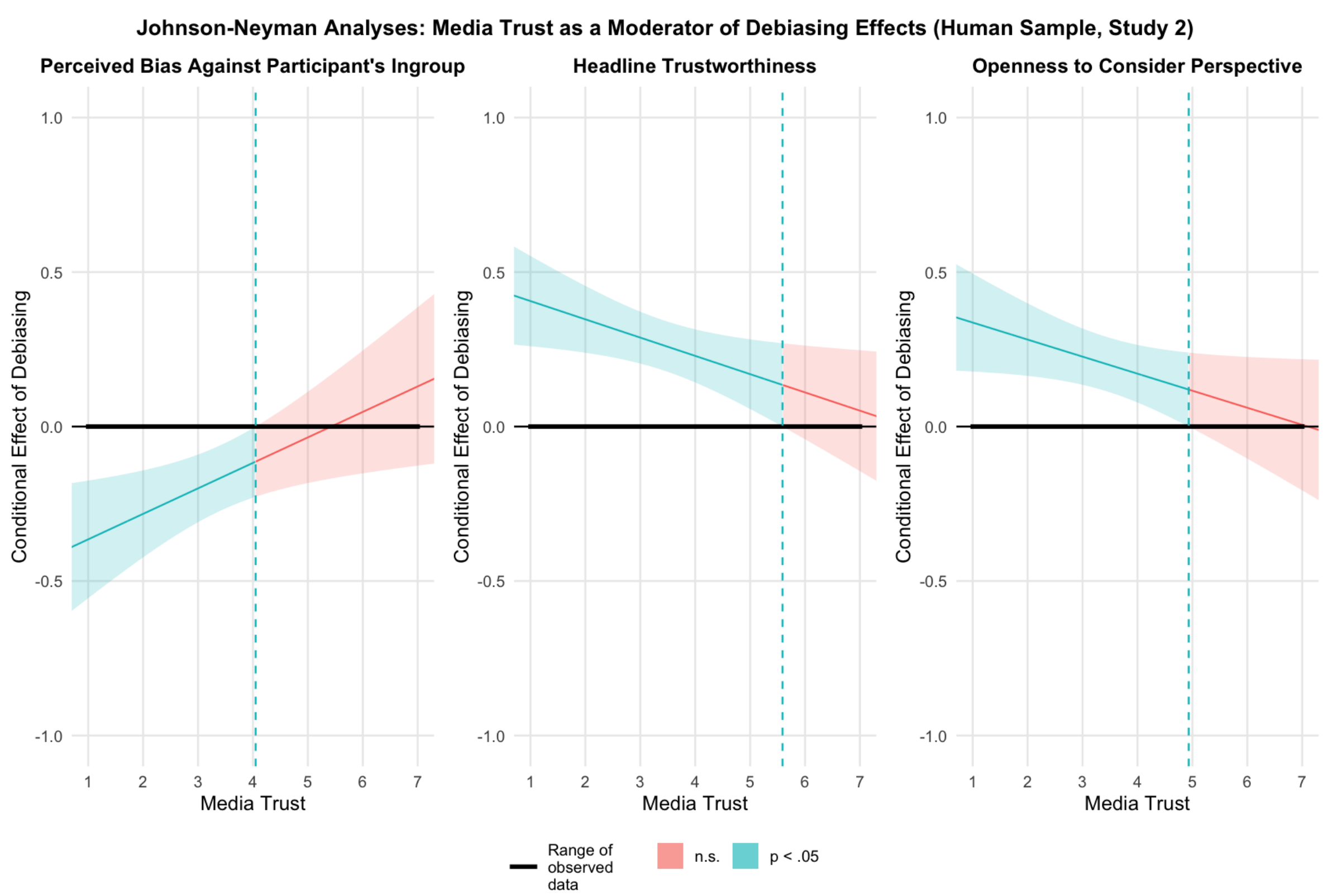

*Note.* Each panel displays the conditional effect of debiasing (original vs. debiased) on the outcome variable as a function of media trust. The solid line represents the point estimate of the conditional effect, and shaded regions indicate 95% confidence bands. Dashed vertical lines mark the Johnson-Neyman thresholds at which the conditional effect crosses the $p$ = .05 boundary. For perceived bias against participants' ingroup, the conditional effect of debiasing was significant below a media trust value of 4.05 and non-significant above it. For headline trustworthiness, the effect was significant below 5.59 and non-significant above it. For openness to consider perspective, the effect was significant below 4.93 and non-significant above it. Blue shading indicates regions where the effect is significant ($p$ < .05); pink shading

indicates non-significant regions. The black horizontal line indicates the range of observed media trust scores. Models include cognitive flexibility and in-group identification as additional moderators and topic familiarity as a covariate. Perceived bias against participants' ingroup was originally measured on a 0–6 scale. n.s. = not statistically significant.

**Three-Way Interactions (Stage 4).** None of the condition × moderator × political group interactions reached significance for any outcome (all *p*s $\geq$ .096), indicating that the moderation patterns described above did not differ between conservative and liberal participants.

### *Silicon Sample Results*

As in Study 1, we conducted the same models with responses from silicon participants generated by o3-mini (see Figure 8).

**Figure 8**

*Model-Adjusted Mean Ratings by Silicon Participants for Original and Debiased Articles, Split by Political Group, Measure in Study 2*

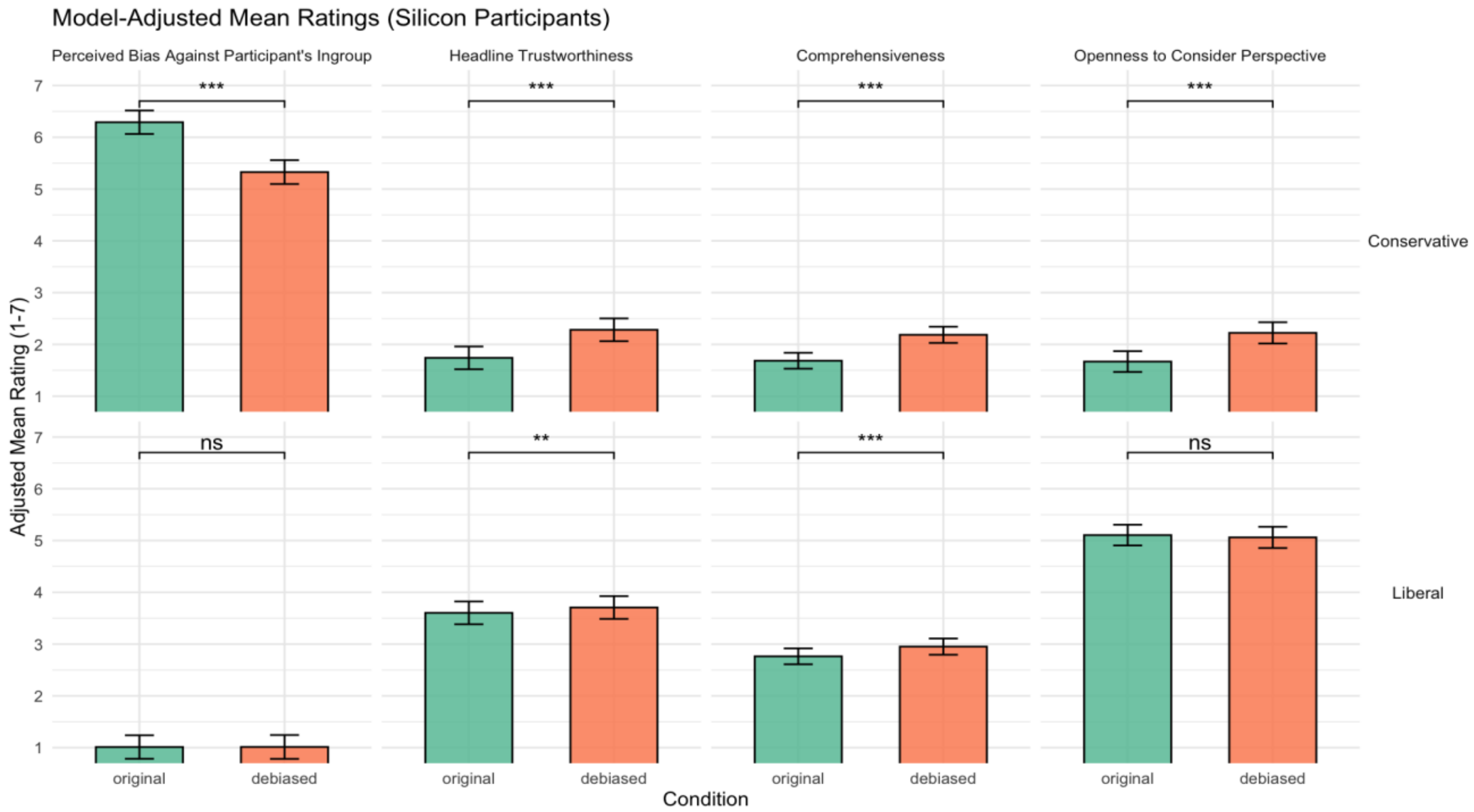

*Note.* Bars represent estimated marginal means from mixed-effects models including condition, political group, their interaction, and topic familiarity as a covariate. Error bars represent 95% confidence intervals. Anti-conservative bias was measured on a 0–6 scale; the remaining outcomes were measured on a 1–7 scale. $**p < .01$, $***p < .001$. n.s. = not statistically significant.

**Confirmatory Tests.** We estimated the same pre-registered Stage 1 and Stage 2 models for the silicon sample.

**Main Effects (Stage 1).** Across all silicon participants, the reframing manipulation significantly affected all four outcomes.

**Anti-Conservative Bias (Manipulation Check**). Debiased headlines were perceived as significantly less biased, $\beta = -0.48$, 95% CI [−0.55, −0.41], $t(401.62) = -13.85$, $p < .001$. Topic familiarity was not significantly associated with perceived bias ($\beta = -0.02$, $p = .205$).

**Headline Trustworthiness.** Debiasing significantly increased perceived trustworthiness, $\beta = 0.32$, 95% CI [0.27, 0.37], $t(330.10) = 12.73$, $p < .001$. Topic familiarity was positively associated with trustworthiness ($\beta = 0.12$, $p < .001$).

**Comprehensiveness.** There was a significant effect of debiasing on perceived comprehensiveness, $\beta = 0.35$, 95% CI [0.30, 0.39], $t(331.25) = 15.68$, $p < .001$. Topic familiarity was positively associated with comprehensiveness ($\beta = 0.10$, $p < .001$).

**Openness to Consider Perspective.** Debiasing significantly predicted greater openness to consider the article's perspective, $\beta = 0.26$, 95% CI [0.20, 0.32], $t(330.57) = 8.59$, $p < .001$. Topic familiarity was positively associated with openness to consider the article's perspective($\beta = 0.11$, $p < .001$).

**Political Group Moderation (Stage 2).** The condition × political group interaction was significant for all four outcomes: perceived bias, $\beta = 0.96$, 95% CI [0.86, 1.07], $t(478.08) = 18.27$, $p < .001$, trustworthiness, $\beta = -0.44$, 95% CI [−0.53, −0.35], $t(3021.08) = -9.62$, $p < .001$, comprehensiveness, $\beta = -0.31$, 95% CI [−0.39, −0.23], $t(336.22) = -7.69$, $p < .001$, and openness to consider the article's perspective, $\beta = -0.60$, 95% CI [−0.70, −0.50], $t(330.68) = -11.50$, $p < .001$.

Simple effects revealed a stark asymmetry, see Figure 8. Among conservative silicon participants, debiased headlines were perceived as significantly less biased (β = −0.96, 95% CI [−1.04, −0.89], $z$ = 25.80, $p$ < .001), more trustworthy (β = 0.54, 95% CI [0.48, 0.61], $z$ = 16.75, $p$ < .001), more comprehensive (β = 0.50, 95% CI [0.44, 0.56], $z$ = 17.34, $p$ < .001), and higher in openness to consider perspective (β = 0.55, 95% CI [0.48, 0.63], $z$ = 15.02, $p$ < .001). Among liberal silicon participants, the effect of debiasing was significant but much smaller for trustworthiness (β = 0.10, 95% CI [0.04, 0.17], $z$ = 3.22, $p$ = .001) and comprehensiveness (β = 0.19, 95% CI [0.13, 0.24], $z$ = 6.55, $p$ < .001), but nonsignificant for perceived bias (β = 0.00, 95% CI [−0.07, 0.07], $z$ = 0.04, $p$ = .971; please note that liberal silicon participants rated MSNBC articles as having virtually no anti-liberal bias, resulting in near-floor scores on this measure; see Figure 8) and openness to consider the article's perspective (β = −0.04, 95% CI [−0.12, 0.03], $z$ = 1.21, $p$ = .225). These effects among conservative silicon participants were substantially larger than the corresponding human effects across all outcomes.

**Exploratory Tests: Moderation Analyses.** We next examined the same moderation models for the silicon sample.

**Moderation by Individual Differences (Stage 3).** In-group identification significantly moderated the effect of debiasing on perceived bias against participant's ingroup, $\beta = -0.14$, 95% CI [−0.21, −0.07], $t(408.36) = -4.08$, $p < .001$, trustworthiness, $\beta = 0.07$, 95% CI [0.02, 0.12], $t(686.63) = 2.62$, $p = .009$, and openness to consider the article's perspective, $\beta = 0.09$, 95% CI [0.03, 0.15], $t(326.92) = 2.81$, $p = .005$, but not comprehensiveness ($\beta = 0.03$, $p = .194$). Media trust significantly moderated the effect of debiasing on perceived bias only, $\beta = 0.10$, 95% CI [0.04, 0.17], $t(407.31) = 3.00$, $p = .003$. Cognitive flexibility did not moderate the effect of debiasing on any outcome (all *p*s > .309).

**Three-Way Interactions (Stage 4).** The condition × media trust × political group interaction was significant for trustworthiness, $\beta = -0.17$, 95% CI [−0.26, −0.08], $t(3079.00) = -3.58$, $p < .001$, comprehensiveness, $\beta = -0.09$, 95% CI [−0.17, −0.01], $t(331.02) = -2.17$, $p = .031$, and perceived bias, $\beta = 0.14$, 95% CI [0.04, 0.25], $t(550.50) = 2.62$, $p = .009$. For trustworthiness and comprehensiveness, the negative coefficients indicate that media trust amplified the debiasing effect more strongly among conservatives than liberals. For bias, the positive coefficient indicates the opposite direction: the moderating role of media trust was stronger among liberals. However, because liberal silicon participants' bias ratings were at floor regardless of condition, this interaction likely reflects negligible variance rather than a substantively meaningful moderation pattern (we therefore do not follow up on the interaction as we do for the other variables next). No other three-way interactions reached significance (all *p*s ≥ .059).

To probe the trustworthiness and comprehensiveness interactions, we conducted Johnson-Neyman analyses separately for conservative and liberal silicon participants (see

Figure 9). For trustworthiness, the effect of debiasing was significant across all observed levels of media trust among conservative silicon participants, and across most of the observed range among liberal silicon participants (non-significant only below a media trust value of 1.68). Among conservative silicon participants, the effect of debiasing increased substantially with media trust: simple slopes ranged from $\beta = 0.39$ ($p < .001$) at low media trust (−1 *SD*) to $\beta = 0.70$ ($p < .001$) at high media trust (+1 *SD*). Among liberal silicon participants, by contrast, the effect was significant but small and essentially constant across media trust levels (βs = 0.11 to 0.14, all *p*s < .02). A similar pattern emerged for comprehensiveness. Among conservatives, the debiasing effect grew with media trust ($\beta = 0.39$ at −1 *SD* to $\beta = 0.62$ at +1 *SD*, all *p*s < .001). Among liberals, the effect was again significant but more stable and smaller in magnitude (βs = 0.16 to 0.24, all *p*s < .001).The three-way interactions thus reflected a difference in the slope of moderation rather than the presence versus absence of an effect: media trust amplified the debiasing effect for conservative silicon participants but had little moderating influence for liberal silicon participants.

**Figure 9**

*Johnson-Neyman Analyses for the Moderating Effect of Media Trust on Debiasing, Split by Political Group and Outcome (Silicon Sample)*

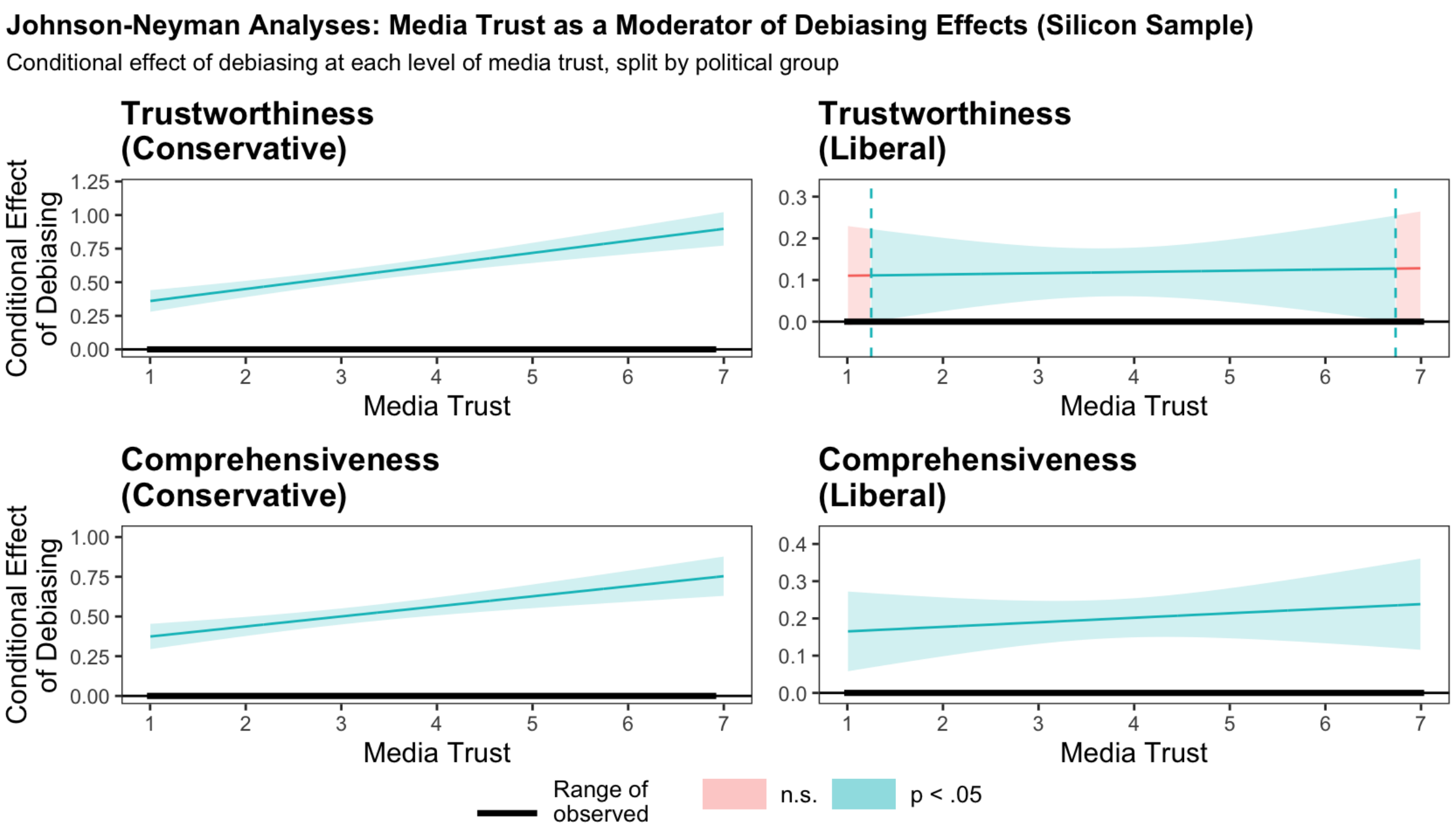

*Note.* Each panel displays the conditional effect of debiasing (original vs. debiased) on the outcome variable as a function of media trust. The solid line represents the point estimate of the conditional effect, and shaded regions indicate 95% confidence bands. Dashed vertical lines mark the Johnson-Neyman thresholds at which the conditional effect crosses the $p = .05$ boundary. Blue shading indicates regions where the effect is significant ($p < .05$); pink shading indicates non-significant regions. The black horizontal line indicates the range of observed media trust scores. For conservative silicon participants (left panels), the debiasing effect was significant across all observed media trust values and increased in magnitude with higher media trust. For liberal silicon participants (right panels), the effect was smaller and relatively constant across media trust levels, with a narrow non-significant region at the low end of the observed range for trustworthiness.

In Study 2 the five additional models again reproduced all four pre-registered effects at $p < .001$, and the planned contrasts show every model overestimating the human debiasing effect on perceived bias (human $\beta = -0.16$ vs. model simple effects of $-0.30$ to $-0.58$), with GPT-5 producing the largest overestimates on three of the four outcomes (Figure 10; Supplementary Section S3). As in Study 1, these per-sample estimates derive from the combined robustness model and may therefore differ slightly from the pre-registered single-sample models reported above.

**Figure 10.**

*Silicon-model robustness, Study 2: debiasing effect (debiased minus original) on each outcome for the human sample and each of the six LLMs, with 95% confidence intervals.*

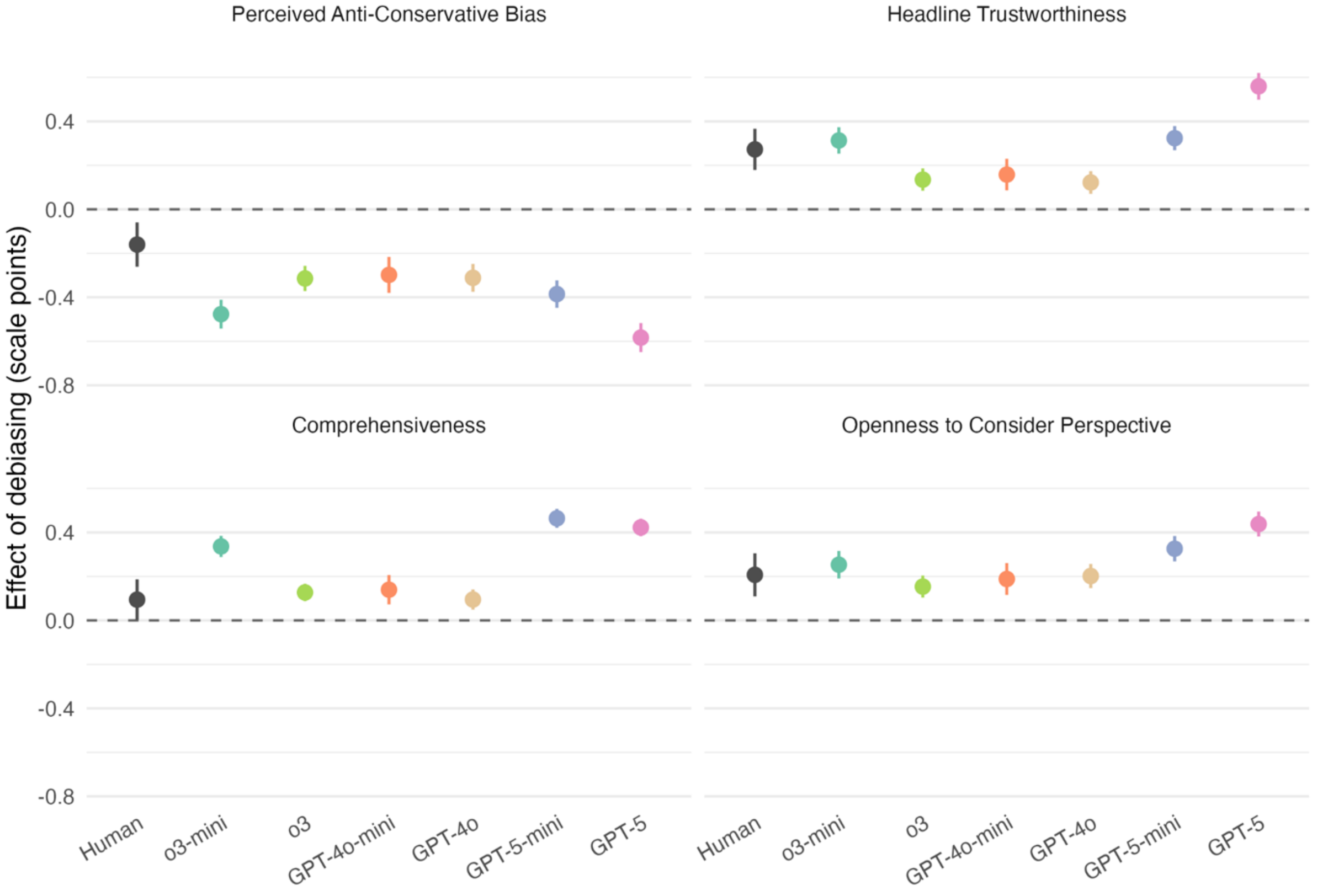

***Comparing Human and Silicon Results***

Due to the inherent difficulty in interpreting four-way interactions (condition × sample type × political group × moderator), we split the analysis into stages of increasing specificity: firs*t* testing condition × sample type, then adding political group, and finally examining moderation within each political group separately. Because media trust was the only moderator that showed significant moderation of the debiasing effect in either the human or silicon within-sample analyses, the cross-sample moderation analyses focused on this variable and controlled for the others.

**Condition × Sample Type.** To test whether the magnitude of the debiasing effect differed between human and silicon participants, we estimated combined models with condition, sample type, their interaction, and familiarity as a covariate. The condition × sample type interaction was significant for perceived bias, $\beta = -0.32$, 95% CI [−0.44, −0.19], $t(650.94) = -4.92$, $p < .001$, and comprehensiveness, $\beta = 0.23$, 95% CI [0.14, 0.33], $t(5079.91) = 4.79$, $p < .001$. It was not significant for trustworthiness, $\beta = 0.04$, 95% CI [−0.06, 0.14], $t(651.62) = 0.75$, $p = .455$, or openness to consider the article's perspective, $\beta = 0.040$, 95% CI [−0.07, 0.15], $t(652.39) = 0.73$, $p = .463$. Effects were larger for silicon participants on bias (silicon: $\beta = -0.48$, 95% CI [−0.57, −0.39], $p < .001$; human: $\beta = -0.17$, 95% CI [−0.26, −0.08], $p < .001$) and comprehensiveness (silicon: $\beta = 0.34$, 95% CI [0.27, 0.41], $p < .001$; human: $\beta = 0.11$, 95% CI [0.04, 0.17], $p = .002$).

**Condition × Sample Type × Political Group.** The three-way interaction was significant for perceived bias, β = 0.53, 95% CI [0.31, 0.76], $t(649.92) = 4.63$, $p < .001$, and openness to consider the article's perspective, β = −0.34, 95% CI [−0.55, −0.14], $t(5128.08) = -3.30$, $p = .001$. It was not significant for trustworthiness, β = −0.10, 95% CI [−0.30, 0.10], $t(649.61) = -1.00$, $p = .316$, or comprehensiveness, β = −0.12, 95% CI [−0.31, 0.07], $t(5897.74) = -1.20$, $p = .231$. Simple effects revealed the source of these interactions. Among conservatives, debiasing was significant in both samples across all four outcomes, but effects were consistently larger for silicon participants (bias: silicon β = −0.96, 95% CI [−1.08, −0.85], $z = 16.69$, $p < .001$ vs. human β = −0.38, 95% CI [−0.49, −0.27], $z = 6.53$, $p < .001$; trustworthiness: silicon β = 0.50, 95% CI [0.40, 0.60], $z = 9.86$, $p < .001$ vs. human β = 0.41, 95% CI [0.31, 0.51], $z = 8.04$, $p < .001$; comprehensiveness: silicon β = 0.48, 95% CI [0.38, 0.57], $z = 9.82$, $p < .001$ vs. human β = 0.19, 95% CI [0.09, 0.28], $z = 3.81$, $p < .001$; openness to consider perspective: silicon β = 0.53, 95% CI [0.43, 0.63], $z = 10.23$, $p < .001$ vs. human β = 0.32, 95% CI [0.22, 0.42], $z = 6.14$, $p < .001$). Among liberals, effects were small or absent in both samples. Neither human nor silicon liberals showed significant effects on bias (human: β = 0.05, 95% CI [−0.07, 0.16], $z = 0.79$, $p = .428$; silicon: β = −0.01, 95% CI [−0.12, 0.11], $z = 0.11$, $p = .915$) or openness to consider the article's perspective (human: β = 0.10, 95% CI [−0.00, 0.20], $z = 1.94$, $p = .053$; silicon: β = −0.03, 95% CI [−0.13, 0.07], $z = 0.55$, $p = .58$). Both liberal samples showed small significant effects on trustworthiness (human: β = 0.14, 95% CI [0.04, 0.24], $z = 2.67$, $p = .008$; silicon: β = 0.13, 95% CI [0.03, 0.23], $z = 2.49$, $p = .013$). On comprehensiveness, silicon liberals showed a significant effect (β = 0.20, 95% CI [0.11, 0.30], $z = 4.19$, $p < .001$) where human liberals did not (β = 0.03, 95% CI [−0.07, 0.12], $z = 0.54$, $p = .591$).

**Moderation by Media Trust Across Samples.** To test whether the moderating role of media trust differed between human and silicon participants, we estimated condition × sample type × media trust models separately within each political group. Among conservatives, the three-way interaction was significant for perceived bias, $\beta = -0.21$, 95% CI [−0.37, −0.05], $t(328.98) = -2.52$, $p = .012$, and trustworthiness, $\beta = 0.16$, 95% CI [0.02, 0.30], $t(318.62) = 2.16$, $p = .031$. It was not significant for comprehensiveness, $\beta = 0.02$, 95% CI [−0.11, 0.15], $t(2467.48) = 0.29$, $p = .774$, or openness to consider the article's perspective, $\beta = 0.07$, 95% CI [−0.08, 0.22], $t(2514.12) = 0.97$, $p = .333$. For bias and trustworthiness, media trust moderated the debiasing effect more strongly for silicon conservatives than for human conservatives, consistent with the within-sample Johnson-Neyman analyses reported above.

Among liberals, the three-way interaction was significant only for openness to consider the article's perspective, $\beta = 0.135$, 95% CI [0.002, 0.267], $t(336.21) = 1.99$, $p = .047$. It was not significant for bias, $\beta = 0.002$, 95% CI [−0.15, 0.15], $t(318.64) = 0.02$, $p = .983$, trustworthiness, $\beta = 0.12$, 95% CI [−0.01, 0.25], $t(327.44) = 1.83$, $p = .068$, or comprehensiveness, $\beta = 0.11$, 95% CI [−0.03, 0.25], $t(2608.18) = 1.61$, $p = .107$. The general absence of significant three-way interactions among liberals is consistent with the small and uniform effects of debiasing observed in this group across both samples.

## Study 2 Discussion

Study 2 tested whether a more substantive reframing intervention could succeed where the minimal lexical substitutions of Study 1 had failed. Indeed, the more substantive intervention significantly shifted conservative readers' trust-relevant judgments across all four outcomes. Conservative participants perceived reframed headlines as less biased, more trustworthy, more comprehensive, and more deserving of consideration. This stands in stark contrast to the uniform null effects observed in Study 1, and suggests that the depth of the intervention matters. Where synonym replacement left the argumentative frame intact and produced no detectable change, reframing the way the issue itself was presented proved sufficient to alter how conservative readers evaluated liberal news content.

The inclusion of liberal participants allowed us to test a possibility that could not be examined in Study 1: whether debiasing produces a backfire effect among the outlet's ideologically aligned audience. No such effect was observed. Liberal participants did not rate reframed headlines less favourably than originals on any outcome. If anything, they showed a small increase in perceived trustworthiness for reframed content. This finding speaks against the concern, raised in the introduction, that neutralizing partisan framing would be perceived by in-group readers as a betrayal (Finkel et al., 2020). Instead, the results suggest that the reframing intervention operated asymmetrically: it improved receptivity among the out-group without degrading the experience for the in-group.

Media trust emerged as the only significant moderator among human participants, attenuating the effect of debiasing on perceived bias, trustworthiness, and openness to consider the article's perspective. In some sense, this pattern may initially be counterintuitive: one might expect participants who trust media more to respond more favourably to reframed content. However, participants with higher media trust may have found the original MSNBC headlines relatively acceptable to begin with, leaving less room for the reframing to improve their judgments (a ceiling-like effect). Participants with lower media trust, by contrast, may have had more negative baseline reactions to the originals, creating greater scope for the reframing to make a difference. This interpretation is consistent with the positive violation-of-expectations account outlined in the introduction (Margoni et al., 2024): readers who anticipate partisan slant may be most responsive when content departs from their negative priors. It is worth noting that this direction of moderation is opposite to the pattern observed in the Study 1 silicon sample, where higher media trust was associated with stronger debiasing effects. Neither cognitive flexibility nor in-group identification statistically significantly moderated the effect of debiasing in the present study, and no three-way interactions reached significance, indicating that the media trust moderation pattern did not differ between conservative and liberal participants.

The silicon sample again showed robust sensitivity to the reframing manipulation. Formal cross-sample comparisons confirmed that the silicon debiasing effect was significantly larger than the human effect for perceived bias and comprehensiveness. For trustworthiness and openness to consider the article's perspective, however, the effects were statistically comparable across samples. This partial convergence represents progress relative to Study 1, where the gap was total, and suggests that when the intervention is substantive enough to produce real human effects, silicon predictions move somewhat closer to human reality on some outcomes. However, among conservative silicon participants, all four effects were highly significant and substantially larger than the corresponding conservative human effects. Among liberal silicon participants, the pattern partially mirrored the human results: significant effects emerged for trustworthiness and comprehensiveness but not for bias or openness to consider the article's perspective. This partial alignment is notable. Where Study 1 revealed a complete disconnect (nulls for humans, strong effects for silicon), Study 2 shows that the silicon sample correctly identified the direction and relative ordering of effects but continued to strongly overestimate their magnitude, consistent with prior findings on LLM effect size inflation (Cui et al., 2025).

The silicon moderation results in Study 2 diverged from the human pattern in informative ways. Whereas media trust was the sole significant moderator for humans, in-group identification was the most consistent moderator for silicon participants, reaching significance for three of four outcomes. This suggests that the model's implicit theory of how individual differences shape responses to debiasing does not align with the actual psychological mechanisms at play. The model appears to overweight the role of partisan identity in determining responsiveness, perhaps because identity-related features are more salient in its training data than the subtler dynamics of media trust. Three-way interactions also emerged in the silicon sample for media trust on trustworthiness and comprehensiveness, indicating that the model's representation of how media trust moderates debiasing effects differs by political group in ways that were not observed among human participants. Cross-sample moderation analyses reinforced this conclusion. Among conservatives, media trust moderated the debiasing effect significantly more steeply for silicon participants than for human participants on bias and trustworthiness. Among liberals, the moderation patterns were largely comparable across samples, consistent with the small and uniform effects in this group. The divergence was thus concentrated where the debiasing effect was largest, precisely the context in which accurate psychological modeling matters most for applied deployment.

Taken together, the results of Study 2 advance the findings of Study 1 in three ways. First, they suggest that the null effects in Study 1 were not due to an inherent imperviousness of conservative readers to debiased content, but rather to the insufficiency of the lexical intervention. When the framing of the news headline itself was altered, conservative readers responded in predicted ways to the debiasing. This supports the interpretation that perceived bias in partisan news is driven primarily by ideological framing rather than by individual word choices (Mohanty et al., 2025), and that effective interventions must operate at the level of how issues are presented, not merely how they are worded. Second, the absence of a backfire effect among liberals suggests that reframing need not come at the cost of alienating the outlet's core audience, an encouraging finding for the practical viability of such interventions. Third, although the human-silicon gap narrowed in Study 2 relative to Study 1 (with comparable effects on trustworthiness and openness to consider the article's perspective), the persistent overestimation on other outcomes and the reliance on a different set of psychological moderators underscore that LLMs cannot yet serve as reliable proxies for predicting how real partisans will respond to editorial interventions. The pattern of silicon effects was directionally consistent with the expectation that reframing would be more effective than lexical substitution, but the model continued to overestimate the magnitude of effects on some outcomes and to rely on a different set of psychological moderators than those that actually mattered for human participants.

## General Discussion

Across two pre-registered experiments, we investigated whether LLM-generated debiasing of partisan news content can shift trust-relevant judgments and information receptivity among ideologically opposed readers, and whether LLMs accurately anticipate the effects of their own interventions. The results yield three principal findings. Table 2 summarizes the pattern of support for each pre-registered hypothesis and results from exploratory moderation across the two studies and the human and silicon samples.

**Table 2**

*Summary of Support for the Pre-Registered Hypotheses and Exploratory Moderations Across Studies and Samples*

| | Study 1 | | Study 2 | |
|---|---|---|---|---|
| | Human | Silicon | Human | Silicon |
| **Confirmatory Tests** | | | | |
| Debiasing reduces perceived anti-conservative bias (manipulation check) | Not supported. | Supported in 6 of 6 models. | Supported. | Supported in 6 of 6 models. |
| Debiasing increases headline trustworthiness (H1a) | Not supported. | Supported in 6 of 6 models. | Supported. | Supported in 6 of 6 models. |
| Debiasing increases perceived headline comprehensiveness (H1b) | Not supported. | Supported in 6 of 6 models. | Supported. | Supported in 6 of 6 models. |
| Debiasing increases openness to consider perspective (H1c) | Not supported. | Supported in 6 of 6 models. | Supported. | Supported in 6 of 6 models. |
| **Exploratory Moderations** | | | | |
| Media trust | All n.s. | Mixed. Media trust was the most frequent moderator but varied by outcome (trust 5/6, bias 3/6, comprehensiveness 2/6, openness 1/6). | Mixed. Debiasing effects on bias, trust, and openness weaker at higher media trust (comprehensiveness n.s.). | Mixed but consistent. Debiasing effects weaker at higher media trust (bias 6/6, openness 5/6, trust 4/6, comprehensiveness 4/6). |
| Cognitive flexibility | Mixed. Only significant for openness : debiasing effect weaker at higher cognitive flexibility. | No consistent moderation. Significant in ≤1 of 6 models per DV (bias 1/6; trust, comprehensiveness, and openness 0/6). | All n.s. | All n.s. |
| Political in-group identification | All n.s. | Largely n.s. Isolated significant interactions in 1 of 6 models each for bias, trust, and comprehensiveness; openness 0/6. | All n.s. | Largely consistent. Debiasing effects stronger at higher in-group identification (trust 6/6, openness 6/6, bias 5/6, comprehensiveness 4/6). |

*Note.* This table presents results at a general level to enhance comprehension. In the Supplementary Materials, we present much more detailed tables that summarize all findings at the level of individual effects (see Supplementary Section S4: Tables S4.1–S4.3 for Study 1 and Tables S4.4–S4.7 for Study 2).

First, the depth of the debiasing intervention is decisive. Study 1 demonstrated that minimal lexical substitution (replacing emotive words with moderate synonyms while preserving narrative structure) had no detectable effect on conservative readers' perceptions of bias, trustworthiness, comprehensiveness, or willingness to engage with the opposing perspective. Study 2 demonstrated that a more substantive reframing intervention, which altered how the issue itself was presented, significantly improved all four outcomes among conservatives. Together, these findings suggest that perceived bias in partisan news is primarily anchored to the ideological frame of an argument (Mohanty et al., 2025) rather than to individual word choices. Surface-level linguistic changes leave the underlying argumentative structure intact and are therefore insufficient to shift how partisan readers evaluate out-group content.

Second, more invasive debiasing did not backfire among the outlet's ideologically aligned audience. As mentioned in Study 2's discussion, a central concern motivating the inclusion of liberal participants in Study 2 was that neutralizing MSNBC's liberal framing may antagonise its core readership (Finkel et al., 2020). This did not occur. Liberal participants did not rate reframed headlines less favourably than originals on any outcome. If anything, the reframing produced a small increase in perceived trustworthiness even among liberals. This asymmetry is encouraging for practical deployment: interventions that improve out-group receptivity without alienating the in-group avoid the zero-sum trade-off that would otherwise limit their viability.

Third, LLMs do not yet possess a sufficiently accurate model of human psychology to serve as autonomous debiasing systems. In Study 1, the disconnect was total: silicon participants showed robust sensitivity to a manipulation that produced no human effects whatsoever. In Study 2, the picture was more nuanced. The silicon debiasing effect was much larger than the human effect for perceived bias and comprehensiveness, but statistically comparable for trustworthiness and openness to consider the article’s perspective. This partial convergence suggests that when the intervention is substantive enough to produce real human effects, the gap between silicon and human responses narrows on some outcomes. Nonetheless, the silicon sample continued to overestimate the magnitude of effects where differences remained, consistent with prior evidence that LLMs overinflate effect sizes when simulating human behaviour (Cui et al., 2025).

The moderation patterns also diverged. For human participants, media trust was the sole consistent moderator, whereas for silicon participants, in-group identification played a more prominent role. Cross-sample moderation analyses confirmed this: among conservatives, media trust moderated the debiasing effect significantly more steeply for silicon participants than for human participants on bias and trustworthiness, whereas among liberals the moderation patterns were largely comparable across samples. This suggests that the model's implicit theory of what drives responsiveness to debiasing does not map onto the psychological mechanisms that actually matter for human partisan readers, and that the misalignment is concentrated precisely where the debiasing effects are largest. These findings extend concerns raised in adjacent domains about LLM-human alignment in political communication (Hackenburg et al., 2025) and underscore the need for human-in-the-loop oversight in any editorial deployment of such systems (Mosqueira-Rey et al., 2022).

Our findings should be read as a proof of concept at the level of headlines and subheadings: LLM debiasing improved cross-partisan receptivity to the unit of text most readers encounter. Whether these effects extend to full-length articles, other outlets across the ideological spectrum, and behavioural rather than evaluative outcomes is an empirical question that any deployment in editorial workflows would first need to answer (we discuss these points in detail in the limitations section).

**Moderation and Individual Differences**

Across both studies, cognitive flexibility and in-group identification did not meaningfully moderate the effect of debiasing among human participants. This null pattern is informative. It speaks against both the prediction that cognitively flexible individuals would be more responsive to neutralized language and the alternative prediction that they would use their cognitive resources for motivated resistance (Kahan & Corbin, 2016). It also speaks against the prediction that strong partisan identifiers would be either especially resistant or especially responsive to debiasing. Instead, where effects existed (Study 2), they were broadly uniform across these individual-difference dimensions.

Media trust was the one moderator that reached significance in the human sample, but only in Study 2 where main effects were present. Among human participants, lower media trust was associated with larger debiasing effects, consistent with a violation-of-expectations account (Margoni et al., 2024). By this violation-of-expectations account, participants who distrust mainstream media expect partisan slant, so reframed content that departs from this expectation produces a larger contrast between prior and stimulus. In contrast, participants with higher media trust, already approach the content with relatively favorable expectations, leaving less room for reframing to shift their judgments. Notably, this direction was opposite to the silicon sample in Study 1, where the debiasing effect on trustworthiness strengthened with higher media trust (interaction $p$ = .050). This reversal is theoretically informative: it suggests that the model's implicit psychology treats trust as a facilitator of receptivity (more trust, more openness to consider the article’s perspective to debiased content), whereas for human readers, it is initial scepticism that creates the room for reframing to make a difference. The opposing moderation directions thus represent a qualitative, not merely quantitative, form of human-silicon misalignment. The model does not simply overestimate effect sizes but misidentifies who will be most affected.

**Limitations and Future Directions**

Several limitations warrant consideration. First, our stimuli consisted of ten opinion headlines and subheadings drawn from a single liberal outlet (MSNBC) and were evaluated by U.S. American participants. This design reflects the outsized role of headlines in contemporary news consumption: an estimated 59% of links shared on Twitter were never clicked (Gabielkov et al., 2016), and roughly 75% of news links on Facebook were forwarded without the sharer clicking on them (Sundar et al., 2024). The design nonetheless carries distinct risks for generalisability. The small stimulus pool means the observed effects may be tied to the specific topics, political figures, and news cycle of April 2025 rather than to partisan opinion content as a general class. Relatedly, our selection strategy deliberately targeted the most divisive headlines from the pretested pool, and the manipulation check confirms that participants experienced the originals as intensely biased. The observed effects are therefore estimated at the high end of bias intensity, where original headlines leave the most room for improvement, and debiasing may produce smaller effects on milder partisan content that provokes less defensive processing at baseline. The restriction to headlines raises a more specific concern. Headlines are short, and a reframed headline may shift judgments precisely because there is no further content against which the new frame can be checked. In full-length articles, where readers have more textual material on which to anchor their judgments and where a debiased frame must be sustained across an entire argument, the same intervention may prove weaker, or may fail when readers detect tension between a neutralised headline and a body that retains its original slant. Our results therefore cannot speak to whether reframing works on articles; they show only that it works on the unit of text most readers encounter.

The single-outlet, single-country, and single survey platform (Prolific) design adds a further constraint. Future research needs to establish whether the pattern holds for conservative outlets (e.g., Fox News) evaluated by liberal readers, since the asymmetry we observed, in which debiasing helped with conservatives without backfiring among liberals, may not survive reversal of the ideological direction. The same applies outside the United States, where audience polarisation is lower and structured differently (Fletcher et al., 2020). Further, although Prolific samples are diverse and were sampled to be quota representative, platform participants are highly familiar with survey research, which could have altered responses relative to more naïve samples drawn from other sources. Practical implementation of AI debiasing in newsroom workflows should not be discussed further until such evidence exists, drawn from larger stimulus pools sampled across outlets, formats, ideological directions, and news cycles.

Second, we selected single-item dependent variables to prevent participant fatigue given the multiple trials participants had to complete. Although these are common in media perception research and were drawn from established scales (Gaziano & McGrath, 1986; Meyer, 1988), multi-item measures would provide more reliable estimates of the underlying constructs. Note, however, that utilizing multi-item measures would likely be prohibitive in a multi-trial design like ours.

A further methodological limitation is the potential for participant fatigue. Evaluating ten consecutive headlines could induce fatigue, which may have introduced measurement error due to increasing inattentiveness over the course of the trials. To mitigate this risk, we deliberately employed single-item outcome measures to keep the task brief and randomized the headline version shown (original versus debiased) both within participants and within articles. Consequently, any fatigue that accumulated over the trials affected both conditions equally in expectation. We ultimately accepted this potential for measurement error because our within-subjects trial structure provides a critical methodological benefit: it allows us to generalize the observed effects across a diverse pool of stimuli, an essential advantage that a simple between-subjects design evaluating a single text could not achieve. Moreover, the possibility of inattention producing less accurate estimates is countered by the high statistical power the design provides.

Yet another limitation concerns the potential that the algorithmic interventions altered the underlying meaning or factual claims of the headlines, rather than exclusively neutralizing partisan slant. While our validation checks involving human raters and semantic embedding models confirmed a high degree of general equivalence between the original and modified texts, a minority of raters still perceived additions, removals, or changes to factual claims (24.0% for the lexical substitutions in Study 1 and 38.1% for the substantive reframing in Study 2). Consequently, we cannot entirely rule out that some semantic alterations occurred alongside the reduction in bias. However, subsequent robustness analyses demonstrated that the primary experimental effects held constant even when statistically controlling for these perceived differences in meaning. This suggests that while minor factual or semantic shifts may have been introduced by the language models, they were not the primary drivers of the observed changes in reader evaluations.

We also relied on self-reported evaluative judgments rather than direct behavioral measures. Although measures of perceived trustworthiness and openness to consider a perspective are essential indicators of early-stage information processing, they do not necessarily translate into overt actions. In real-world news consumption, a reader's self-reported willingness to engage with a headline may not predict whether they actually click on the article, read it to completion, or share it within their social networks. Crucially, this reliance on self-report introduces the danger of demand characteristics. Participants, particularly in the substantive reframing condition of the second experiment, might have deduced the underlying aim to neutralize political slant and adjusted their responses to align with perceived expectations rather than their genuine evaluations. However, given the subtle manipulation (and that each trial was independently manipulated; participants never saw both versions of the same news headline), we consider the risk of demand characteristics low. Nevertheless, unobtrusive behavioral tracking, such as measuring dwell time, click-through rates, or sharing decisions in a simulated social media environment is generally more robust against such experimental artifacts. Future research should incorporate these behavioral metrics to determine whether the psychological shifts observed in the present studies translate into meaningful changes in how citizens navigate polarized information spaces.

The debiasing interventions and silicon participants in both studies were generated with a single model, OpenAI's o3-mini. LLMs differ in architecture, training data, and alignment procedures, and a misalignment pattern observed in one model could be an idiosyncrasy of that model's training rather than a general property of current systems. Future work should estimate whether more recent models perform differently in linguistic debiasing of text. Regarding our central methodological conclusion that silicon participants cannot substitute for human samples in this domain, we conducted the silicon participant analyses with five additional models from the GPT family varying in scale and generation: GPT-4o mini, GPT-4o, o3, GPT-5 mini, and GPT-5. The central misalignment replicated across all six models: every model showed robust debiasing effects in Study 1, where human effects were uniformly null, and every model overestimated the effect of reframing on conservatives' perceived bias in Study 2. The in-group identification moderation absent in the human data emerged in all six models in Study 2, and three models predicted a backfire effect among liberal readers that did not occur in the human sample. Misalignment did not diminish with model scale or generation; strikingly, GPT-5 produced the largest overestimates on several Study 2 outcomes. Nonetheless, all six models are from a single provider, and models trained with different architectures or reinforcement-learning protocols remain untested, meaning our findings could reflect model-specific artifacts rather than a universal limitation of artificial intelligence.

Furthermore, the capabilities of large language models are evolving rapidly. The absence of improvement from o3-mini to GPT-5 suggests that misalignment is not simply an artifact of model scale, but future iterations or models explicitly fine-tuned on psychological data may perform differently. Continuous benchmarking will therefore be necessary to determine whether these simulation deficits are an inherent feature of such models or a transient characteristic of their current development phase.

The silicon pipeline held decoding settings constant at each model's API default: no temperature parameter was passed in any call. For the reasoning models (o3-mini, o3, GPT-5, GPT-5 mini), the interface exposes no temperature parameter; for the GPT-4o models, which do accept one, the default (temperature = 1.0) was deliberately left in place so that no model received tuning unavailable to the others. Temperature affects only how much a model's answers vary randomly from one response to the next. Because each silicon participant was randomly assigned which headline version to rate, this random variation cannot systematically favor one condition over the other: a higher temperature could have made real effects harder to detect, but it could not have created the effects we observe. Consistent with this, the GPT-4o models, run at the default temperature, showed the same effects in the same direction and at the same significance levels as the reasoning models, for which temperature cannot be set at all. Temperature settings are therefore unlikely to explain the misalignment pattern we report. We cannot rule out that effect magnitudes would shift somewhat under other settings, however, and future silicon-sample work could test this directly by varying temperature where the interface permits.

Finally, our outcomes capture the earliest stage of information processing and nothing beyond it. We measured perceived trustworthiness, perceived comprehensiveness, and openness to consider the presented perspective. These track whether readers will engage with outgroup content versus dismiss it before evaluating it. We did not track belief change. A conservative who rates a reframed MSNBC headline as more trustworthy has not thereby moved on the underlying issue. Our intervention targets receptivity rather than persuasion, the outcome that moral reframing research addresses directly (Feinberg & Willer, 2019), and the two can dissociate. This scope also qualifies how the term bias should be read here. Our manipulation reduced the partisan framing cues that signal outgroup authorship and provoke defensive processing (Walker et al., 2025). Reducing those cues is distinct from producing an objectively more accurate or less slanted account, and our data speak only to the former. Whether the resulting rise in trust toward incongruent content is epistemically desirable therefore depends on two further conditions. The first concerns which good one applies as the standard. Cross-partisan flow of information is one such good: deliberative democracy requires citizens to encounter and weigh views they do not already hold (Mutz, 2006), so moving readers from dismissal to consideration has value on its own. Attitude change and compromise constitute a second, which requires readers to revise positions once considered. Our studies address the first and leave the second untested. Consideration may rise while beliefs stay fixed, since closer engagement with opposing content can prompt counter-arguing rather than updating (Taber & Lodge, 2006). The second condition concerns the factual quality of the content itself. To ensure ecological validity, we sampled headlines as they naturally appeared on the MSNBC platform and did not select them on the basis of veracity, so we could no*t* test whether the manipulation produces different effects for accurate and inaccurate content. Receptivity gains are epistemically valuable when they attach to accurate reporting, since citizens then engage with true information they might otherwise dismiss for partisan reasons.

The same method applied to false or misleading content would increase receptivity to misinformation. Debiasing is therefore dual-use: it lowers defensive processing toward outgroup content regardless of whether that content is accurate, and deployment would need to couple reframing with verification so that the intervention is applied only to content already checked. Future work should accordingly do two things: stratify stimuli by official fact-check results, and track outcomes downstream, testing whether sustained exposure to reframed and verified material changes what readers believe, share, and read in full, rather than only whether they are willing to hear it.

## Conclusion

The present research demonstrates that LLM-generated debiasing of partisan news content can improve cross-partisan receptivity, but only when interventions engage with the ideological framing of the content rather than merely softening its language. At the same time, LLMs systematically overestimate the effectiveness of their own interventions and seem to rely on a different set of psychological assumptions than those governing actual human responses. As newsrooms and platforms are integrating generative AI into editorial workflows (Cools & Diakopoulos, 2024), our findings make clear that human oversight remains essential. Models can be powerful tools for generating candidate interventions, but they cannot yet be trusted to evaluate whether those interventions will work.

**AI disclosure.** Apart from the silicon participants generated as part of the substantive contribution of this paper, AI tools (Claude, Gemini) were used to assist in producing analysis scripts and in language editing. We take full responsibility for the content.